\documentclass[journal]{IEEEtran}

\usepackage{cite}
\usepackage{graphicx}
\usepackage{amsmath,amssymb}
\usepackage{algorithm,algpseudocode}
\usepackage{color}
\usepackage{url}
\usepackage{bm}
\usepackage{siunitx}
\usepackage{caption}
\usepackage{enumitem}
\usepackage{subcaption}
\usepackage{fancyhdr}
\usepackage{booktabs}
\usepackage{multirow}
\usepackage[table]{xcolor}
\usepackage{soul}
\usepackage{comment}
\usepackage{hyperref}

\usepackage{dblfloatfix}

\usepackage{xspace}
\newcommand{\ie}{i.\,e.,\xspace}
\newcommand{\eg}{e.\,g.,\xspace}

\title{Diffusion-Based Impedance Learning for Contact-Rich Manipulation Tasks}

\author{
  Noah Geiger,
  Tamim Asfour,\thanks{N. Geiger and T. Asfour are with the Institute for Anthropomatics and Robotics, Karlsruhe Institute of Technology, Karlsruhe, Germany; e-mail: noah.geiger@student.kit.edu, asfour@kit.edu}
  Neville Hogan\thanks{N. Geiger, N. Hogan, and J. Lachner are with the Department of Mechanical Engineering, Massachusetts Institute of Technology, Cambridge, USA; N. Hogan, and J. Lachner are also with the and Department of Brain and Cognitive Sciences, Massachusetts Institute of Technology, Cambridge, USA; e-mail: ngeiger@mit.edu, neville@mit.edu, jlachner@mit.edu}
  and~Johannes Lachner
 }

\begin{document}

\maketitle

\begin{abstract}
Learning-based methods excel at robot motion generation but remain limited in contact-rich physical interaction. Impedance control provides stable and safe contact behavior but requires task-specific tuning of stiffness and damping parameters. We present Diffusion-Based Impedance Learning, a framework that bridges these paradigms by combining generative modeling with energy-consistent impedance control. A Transformer-based Diffusion Model, conditioned via cross-attention on measured external wrenches, reconstructs simulated Zero-Force Trajectories (sZFTs) that represent contact-consistent equilibrium behavior. A SLERP-based quaternion noise scheduler preserves geometric consistency for rotations on the unit sphere. The reconstructed sZFT is used by an energy-based estimator to adapt impedance online through directional stiffness and damping modulation. Trained on parkour and robot-assisted therapy demonstrations collected via Apple Vision Pro teleoperation, the model achieves sub-millimeter positional and sub-degree rotational accuracy using only tens of thousands of samples. Deployed in real-time torque control on a KUKA LBR iiwa, the approach enables smooth obstacle traversal and generalizes to unseen tasks, achieving 100\% success in multi-geometry peg-in-hole insertion. The code for all experiments is publicly available on \href{https://github.com/StrokeAIRobotics/Diffusion-Based-Impedance-Learning_T-RO}{GitHub} and videos of the experiments are available on the \href{https://strokeairobotics.github.io/Diffusion-Based-Impedance-Learning_T-RO/}{project website}.

\end{abstract}


\section{Introduction}
Robotic behavior arises from the interaction of two fundamentally different domains. On the one hand, motion planning and learning algorithms are based on data processing and operate on symbolic or numerical representations of virtual desired behavior, such as trajectories, poses, or task objectives. These representations are processed unidirectionally and are constrained only by temporal causality (``no output before input'') and boundedness. They are not subject to physical laws. In this paper, we refer to this as the \textit{information domain}~\cite{hogan2017physical}.

Physical interaction with the environment, on the other hand, is governed by bidirectional behavior. It is constrained by physical laws like energy conservation, entropy production, gravity, etc. In this paper, we refer to this as the \textit{energy domain}~\cite{hogan2017physical}. Ignoring these physical constraints will yield  robot control problems with stability (passivity)~\cite{lachner_shaping_2022, lachner2025_stiffnessFormulation} and safety~\cite{lachner_energy_2021}. 

Robots operating in contact-rich and unstructured environments must bridge these two domains: desired motions must be inferred and represented in the information domain, yet executed through control algorithms that take into account  energy exchange with the environment. This requirement is critical in assembly, robot-assisted rehabilitation, and other contact-rich tasks where visual feedback may be limited and task success depends on regulating physical interaction rather than simply following a path.

The relation between motion planning (information domain) and physical interaction (energy domain) is well described by mechanical impedance~\cite{hogan1987stable,lachner2022geometric,stramigioli_energy-aware_2015}. Impedance Control has been widely adopted to guarantee stability \cite{hogan1987stable} and safety \cite{lachner_energy_2021} for robots that go in and out of contact. The separation of both domains can be formalized by equivalent networks, a generalization
of the equivalent-circuit concept~\cite{helmholtz1853distribution}, to represent nonlinear dynamics and arbitrarily complex physical networks as a combination of primitives~\cite{hogan2014general, nah2023, nah2024}.

While Impedance Control provides a powerful framework for shaping robot–environment interaction~\cite{hogan2014general}, its effectiveness critically depends on the choice of stiffness and damping parameters~\cite{hogan1985impedance}. Too much stiffness can lead to jamming, while too little can prevent task execution. Selecting these parameters is typically an iterative and time-consuming process, as current physics-based simulation tools are not yet sufficiently accurate or reliable to support fully offline tuning \cite{Collins_2019}. 

Recent advances in contact-rich manipulation can be divided into three main strands: 1) \textit{model-based approaches},  2) \textit{statistical-impedance heuristics}, and 3) \textit{learning-based approaches}. 

Model-based approaches incorporate explicit contact dynamics including friction cones, complementarity constraints, and mixed-integer formulations \cite{posa2014direct,manchester2020variational,pang2023global,marcucci2017approximate,hogan2020reactive,anitescu1997formulating,aydinoglu2022real,suh2022bundled,suh2025dexterouscontact, Li2018, Li2025, shaw2022, ANAND2023, Youssef2024}. 
These methods produce behavior consistent with physical contact but face challenges: high computational cost, parameter sensitivity, and limited robustness in unstructured settings. Most approaches require task-specific tuning of parameters. The selected control gains typically do not generalize across environments and tasks. Consequently, these approaches are not ideal for unstructured environments with unknown contact dynamics. These limitations motivate the development of learning-based frameworks capable of inferring and adapting stiffness online from observed interaction data.

Between explicit physics-based models and fully learning-based policies, a class of statistical-impedance heuristics have emerged within the learning-from-demonstration literature. These approaches typically model demonstrated motions using probabilistic representations such as Gaussian Mixture Models, where the mean trajectory defines the nominal motion and the covariance is interpreted as a measure of task relevance or confidence~\cite{Abi-Farraj_2017, Najafi_2017}. Impedance parameters are then modulated heuristically based on this statistical variability, with higher stiffness assigned to directions of low variance. These approaches have shown to be effective for structured tasks with consistent geometry. However, these methods are still limited in contact-rich settings where changing environmental constraints are not reflected in the statistical variability of the demonstrations.

Learning-based approaches, in contrast, primarily operate in the information domain. Reinforcement learning and sampling-based optimization methods~\cite{theodorou2010generalized, kalakrishnan2011stomp, stulp2012reinforcement, williams2017model, Yang2022, tahmaz2025} can achieve context-aware behavior, but typically lack an explicit representation of impedance, which prescribes a bidirectional relationship between motion and generalized force~\cite{Hogan1985}. Instead, these methods rely on reward functions that penalize either force or position deviations. As a result, they tend to converge toward admittance-like behaviors that suppress interaction forces or motion deviations, rather than explicitly regulating impedance.

More recent generative approaches, including diffusion policies~\cite{Chi_2023_Diffusion}, flow-matching policies~\cite{braun2024, zhang2025flow}, and Transformer-based policy models~\cite{Shridhar2022, Kim2024}, have demonstrated impressive performance in motion generation across tasks and embodiments~\cite{trilbmteam2025largebehavior}. However, these methods primarily focus on trajectory generation and are typically executed through PD controllers with fixed proportional and derivative gains to maintain stability during contact. Compliance in such systems is therefore not explicitly controlled but emerges indirectly through low control gains. While this strategy can prevent hardware damage, it provides limited control over physical interaction behavior in contact-rich or safety-critical scenarios where deliberate impedance modulation is essential.

This paper addresses this gap by formulating impedance adaptation as an equilibrium reconstruction problem rather than a trajectory learning problem. We present \emph{Diffusion-Based Impedance Learning}, a framework that integrates generative modeling with energy-consistent impedance control. While the robot moves on a fixed nominal Zero-Force Trajectory (ZFT)\footnote{The Zero-Force Trajectory (ZFT), introduced by Hogan \cite{hogan1985impedance}, refers to the commanded equilibrium motion in the unconstrained case: the unique end-effector pose at which the interaction wrench vanishes.}, a conditional Transformer-based Diffusion Model reconstructs a simulated Zero-Force Trajectory (sZFT) from contact-perturbed motion and measured interaction wrenches. The reconstructed sZFT represents an inferred equilibrium that is consistent with the observed physical interaction and serves as the reference for impedance modulation. A directional adaptation scheme selectively reduces stiffness along non-task-relevant directions while preserving rigidity along directions essential for task execution. 

Other work, in addition to that presented here, has also begun to combine generative models with compliant control for contact-rich manipulation. The Adaptive Compliance Policy (ACP)~\cite{hou2025adaptive} shares several conceptual similarities with the present work, but differs in sensing modality, demonstration paradigm, controller architecture, and the role of the inferred equilibrium. ACP combines visual observations, force sensing, and compliant kinesthetic teaching to directly predict compliance-related quantities for a low-level controller. In contrast, our approach operates solely on robot kinematics and measured interaction wrenches, reconstructing a contact-consistent equilibrium trajectory from teleoperated demonstrations and deriving impedance adaptation through an energy-based estimator. Conceptually, both ACP’s Virtual Target Pose and the proposed equilibrium reconstruction represent interaction-conditioned equilibria. However, the former is directly tracked as a force-generating target, whereas the latter serves as an intermediate representation for estimating impedance adaptation while the robot continues to follow its nominal motion reference.

In this way, our approach bridges data-driven learning and energy-consistent control: Diffusion Models explain physical interaction by reconstructing contact-consistent equilibria, while impedance control enforces stability and physical consistency during execution. Our method operates purely on robot kinematics and measured interaction wrenches, without relying on visual observations. 

\noindent
We validated the framework on a KUKA LBR iiwa in two contact-rich scenarios: parkour-style obstacle traversal and multi-geometry peg-in-hole insertion. Training data were collected through teleoperation using Apple Vision Pro (AVP) equipped with a markerless pose-tracking framework~\cite{park2024} integrated into the robot controller. Both tasks expose the limitations of fixed impedance and simple adaptation schemes, which require careful, task-specific parameter tuning and still tend to jam at obstacles or fail to achieve insertion. In contrast, Diffusion-Based Impedance Learning achieves consistent success without manual retuning. 

This result is particularly noteworthy in the peg-in-hole experiments, as the Diffusion Model was not trained on this task. Despite the absence of peg-in-hole demonstrations during training, the system achieved a 100\% success rate across all peg geometries. This generalization was obtained using only ~1.6 hours of teleoperated training data (72,000 samples). The results summarized in Table~\ref{tab:preview_results} underscore the central contribution of this work: robust contact-rich manipulation in unstructured environments requires explicitly bridging learning in the information domain with control in the energy domain.

\begin{table*}[!t]
    \centering
    \caption{Preview of experimental outcomes (full details in Section~\ref{sec:experiments}). 
    Fixed impedance fails in contact-rich tasks, while Diffusion-Based Impedance Learning achieves robust success by reconstructing a contact-consistent sZFT.}
    \vspace{0.5em}
    \begin{tabular}{lcccc}
        \toprule
        \textbf{Controller} & \textbf{Parkour} & \textbf{Cylindrical Peg} & \textbf{Square Peg} & \textbf{Star Peg} \\
        \midrule
        Fixed impedance & $\times$ (stopped at first obstacle) & 100\% & 13\% & 0\% \\
        \rowcolor{gray!15} Diffusion-Based Impedance Learning & \checkmark~(smooth traversal) & 100\% & 100\% & 100\% \\
        \bottomrule
    \end{tabular}
    \label{tab:preview_results}
\end{table*}

The code for the Transformer-based Diffusion Model, the robot controller, 
and the Apple Vision Pro teleoperation framework is available in our 
GitHub repository. Demonstration videos of all 
experiments can be found on the project website.

\section{Preliminaries}
First, we outline the process in both domains: Transformer-based diffusion in the information domain and energy-based Impedance Control in the energy domain. 

\subsection{Diffusion Models}

\begin{figure*}[!b]
    \centering    \includegraphics[width=0.7\textwidth, trim={0 0.5cm 0 0}]{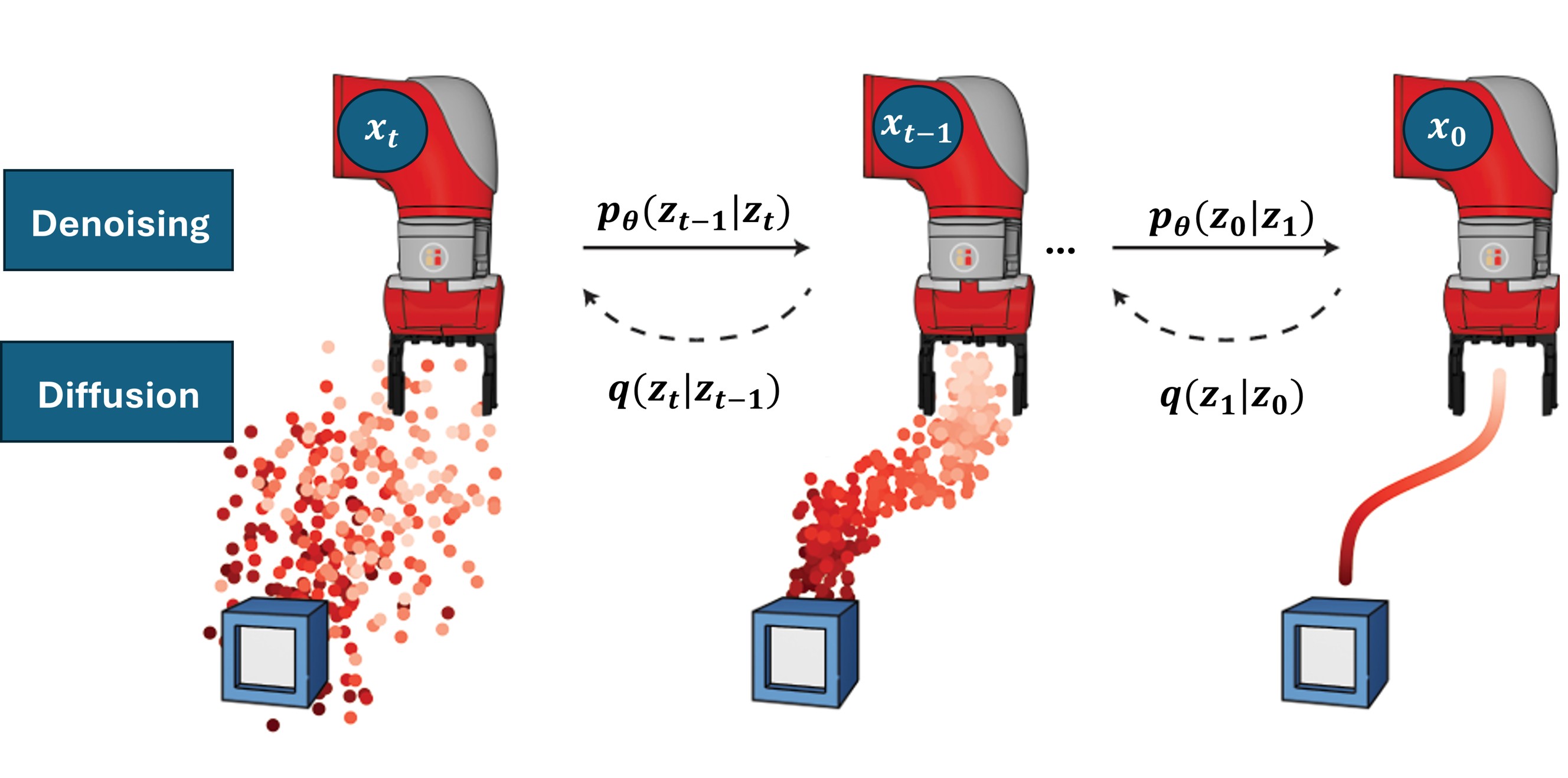}
    \caption{Diffusion overview. Bottom: dashed arrows depict the forward process $q(z_t\!\mid\!z_{t-1})$ which gradually corrupts a clean trajectory $z_0$. Top: solid arrows depict the reverse process $p_\theta(z_{t-1}\!\mid\!z_t)$ which is learned to reconstruct structured robotic behavior. Figure adapted from \cite{Janner2022}.}
    \label{fig:diffusion_overview}
\end{figure*}

Diffusion Models learn to sample from a data distribution by approximating the reverse-time dynamics of a Gaussian diffusion process~\cite{Janner2022}. 
Let \(\{z_i\}_{i=1}^N\) be independent and identically distributed samples from the data distribution \(p_{\text{data}}\); the goal is to learn a parametric family of distributions \(\{p_\theta\}_{\theta\in\Theta}\) such that \(p_\theta \approx p_{\text{data}}\).
A forward Markov chain maps a clean sample \(z_0\) to (approximately) \(\mathcal N(0,\mathbf I)\) over \(T\) steps. The learned reverse-time chain iteratively denoises \(z_t\) to produce a sample that approximates \(p_{\text{data}}\) (Figure~\ref{fig:diffusion_overview}).

\subsubsection{Forward Diffusion}
The forward diffusion process defines a Markov chain with Gaussian transitions
\begin{equation}
    q(z_t \mid z_{t-1}) \;=\; \mathcal{N}\!\left(\sqrt{\alpha_t}\, z_{t-1}, \, (1-\alpha_t)\,\mathbf{I}\right),
\end{equation}
where $\alpha_t := 1-\beta_t$ and $\{\beta_t\}_{t=1}^T \subset (0,1)$ is a predefined noise schedule. The quantity $(1-\alpha_t)=\beta_t$ is the per\mbox{-}step diffusion variance, which controls the signal-to-noise ratio across time. After $T$ steps with $\bar\alpha_T \!:=\! \prod_{i=1}^T \alpha_i \approx 0$, the marginal distribution approaches a standard normal distribution , $z_T \approx \mathcal{N}(0,\mathbf{I})$. The marginal distribution at step $t$ has the closed form
\begin{equation}
    q(z_t \mid z_0) \;=\; \mathcal{N}\!\big(\sqrt{\bar\alpha_t}\, z_0,\; (1-\bar\alpha_t)\,\mathbf{I}\big),
\end{equation}
which admits the equivalent reparameterization
\begin{equation}
    z_t \;=\; \sqrt{\bar\alpha_t}\, z_0 \;+\; \sqrt{1-\bar\alpha_t}\,\varepsilon, 
    \qquad \varepsilon \sim \mathcal{N}(0,\mathbf{I}).
\end{equation}

\subsubsection{Noise Schedule}
The per-step variances $\{\beta_t\}_{t=1}^T$  define the noise schedule, which controls how the signal-to-noise ratio $\mathrm{SNR}_t=\bar\alpha_t/(1-\bar\alpha_t)$ decays from $\infty$ to $0$ over time. A well-chosen schedule balances the complexity of denoising, keeping early steps near-deterministic and later steps close to pure noise. \cite{Nichol2021, Ho2020}

\subsubsection{Reverse Process and Learning Objective}
The true reverse $q(z_{t-1}\!\mid\!z_t)$ is intractable, as it marginalizes over $z_0$. However, since the forward chain is linear–Gaussian, the posterior $q(z_{t-1}\!\mid\!z_t,z_0)$ is itself Gaussian with mean $\tilde\mu_t(z_t,z_0)$ and variance $\tilde\beta_t\mathbf I$ \cite{SohlDickstein2015}. The model represents $p_\theta(z_{t-1}\!\mid\!z_t)=\mathcal N(\mu_\theta(z_t,t),\Sigma_\theta(z_t,t))$ and fit $\theta$ by minimizing the variational bound $\sum_t D_{\mathrm{KL}}\!\big(q(z_{t-1}\!\mid\!z_t,z_0)\,\|\,p_\theta(z_{t-1}\!\mid\!z_t)\big)$. With the $\varepsilon$-parameterization,
\[
\mu_\theta(z_t,t)=\frac{1}{\sqrt{\alpha_t}}\!\left(z_t-\frac{\beta_t}{\sqrt{1-\bar\alpha_t}}\,\varepsilon_\theta(z_t,t)\right),
\]
and $z_t=\sqrt{\bar\alpha_t}z_0+\sqrt{1-\bar\alpha_t}\,\varepsilon$, the objective reduces (up to constants) to the noise-prediction loss \cite{Ho2020}:
\[
\mathcal L_{\mathrm{simple}}(\theta)=\mathbb E_{t,z_0,\varepsilon}\!\left[\;\big\|\varepsilon-\varepsilon_\theta\!\left(\sqrt{\bar\alpha_t}z_0+\sqrt{1-\bar\alpha_t}\,\varepsilon,\,t\right)\big\|_2^2\right].
\]

\subsubsection{Training and Inference Algorithms}
Training follows a noise-prediction objective: at each iteration, a clean sample $z_0\!\sim\!p_{\text{data}}$, a timestep $t\!\in\!\{1,\dots,T\}$ (uniform), and Gaussian noise $\varepsilon\!\sim\!\mathcal N(0,\mathbf I)$ are sampled. The closed-form forward diffusion produces $z_t$, then the network $\varepsilon_\theta$ is updated to minimize mean-squared error between $\varepsilon$ and its prediction $\varepsilon_\theta(z_t,t)$ \cite{Ho2020,Holderrieth2025}. At inference, generation starts from $z_T\!\sim\!\mathcal N(0,\mathbf I)$ and applies the learned reverse transitions $p_\theta(z_{t-1}\!\mid\!z_t)$ for $t{=}T,\dots,1$, yielding $z_0$ as a sample from the model \cite{Ho2020}. For robotic trajectory generation, this iterative denoising maps pure noise into trajectories that are consistent with task goals (Figure~\ref{fig:diffusion_overview}).

\subsubsection{Conditional Diffusion}
To generate task-consistent behavior, the denoising process is conditioned on an additional context variable $c$ (\eg goals, scene state, or interaction wrench). The reverse model and noise predictor then become $p_\theta(z_{t-1}\!\mid\!z_t,c)$ and $\varepsilon_\theta(z_t,t,c)$. 

Conditioning can be implemented in several ways, including (i) early fusion (\eg feature concatenation), (ii) cross-attention over context tokens for Transformer denoisers, or (iii) lightweight fusion layers \cite{Liu2025,Mishra2023}.

In our approach, conditional diffusion is central. By conditioning on external wrenches, the model reconstructs sZFTs that are consistent with physical interaction. This ties the denoising process in the information domain to the regulation of interaction in the energy domain.
More details about Diffusion Models can be found in \cite{song2020denoising, Ho2020, Wolf2025, Holderrieth2025}.

\subsubsection{Transformer-based Diffusion Model}
In this paper, we choose a Transformer-based denoising network since robotic trajectories are time series with context-dependent effects (\eg contact events with delayed consequences). Self-attention provides a content-based global receptive field and fully parallelizable training, which is advantageous over Recurrent or Convolutional Neural Networks for sequence modeling \cite{Vaswani2017, Kim2024}.

At each diffusion step, pose tokens are formed from translations and unit-quaternions. These tokens are embedded and enriched with learned positional and timestep embeddings. Interaction signals (forces and moments) are projected to context tokens and integrated through a multi-head cross-attention layer. Here, trajectory tokens serve as queries while context tokens serve as keys and values. The combined sequence is then processed by stacked self-attention and feed-forward layers, and a lightweight Multi-layer Perceptron (MLP) head predicts the 7D noise (3D translations and 4D unit-quaternions) for each token. This architecture captures temporal structure while conditioning on interaction signals, a requirement that is central for contact-rich applications \cite{Vaswani2017, Liang2024, Dosovitskiy2021, Jaegle2021, Wang2022, Lin2022}.
More details about Transformer-based models can be found in \cite{Vaswani2017, Wang2022, Lin2022, Wolf2025}.

\subsection{Energy-based Impedance Control}\label{subsec:IC}
Impedance Control regulates the dynamic relationship between motion and interaction wrenches. Rather than commanding exact positions, the robot behaves like a virtual mechanical system characterized by stiffness, damping, and inertia\footnote{And possibly higher-order dynamic effects.}. In contrast to admittance control\footnote{Admittance control requires additional methods to guarantee passivity \cite{Kramberger2018}.}, 
the impedance formulation enables the robot to maintain compliance while ensuring stability when interacting with passive environments \cite{Hogan1985, Hogan2014, Hogan2015, lachner2022geometric}.

From an energy perspective, Impedance Control transforms the potential energy stored in a virtual spring to the work done by external wrenches \cite{Lachner2021}. 
Let the pose displacement be 
\begin{subequations}
\begin{align}
\Delta \mathbf{x} &=
\begin{pmatrix}
\mathbf e_t \\[2pt] \mathbf e_r
\end{pmatrix}
\in \mathfrak{se}(3), \\[4pt]
\mathbf e_t &:= \bm{p} - \bm{p}_0 \in \mathbb R^3, \\[4pt]
\mathbf e_r &:= \log\!\big( \bm{Q}_0 \bm{Q}^{-1}\big) \in \mathbb R^3 \simeq \mathfrak{so}(3).
\end{align}
\end{subequations}
where the observed pose is $(\bm{p},\bm{Q})\in \mathbb R^3 \times \mathbb S^3$ and the equilibrium pose is $(\bm{p}_0,\bm{Q}_0)\in \mathbb R^3 \times \mathbb S^3$.
Any unit quaternion admits the unit axis–angle representation
\begin{equation}\label{eq:unitAxisAngle}
    \bm{Q} = \big( \cos(\tfrac{\theta}{2}),\; \mathbf u \sin(\tfrac{\theta}{2}) \big),
    \quad \|\mathbf u\|=1,\; \theta \in [0,\pi].   
\end{equation}
In this formulation, the logarithm map is applied to the error quaternion $\bm{Q}_0 \bm{Q}^{-1}$ to return the axis–angle error vector $\bm{e}_r = \bm{u}_0 \ \theta_0$.

Let $\bm{K}_t,\bm{K}_r\in\mathbb R^{3\times 3}$ be diagonal, positive semidefinite translational and rotational stiffness matrices and $\bm{K}=\mathrm{diag}(\bm{K}_t,\bm{K}_r)$.
Define the potential energy
\begin{equation}
\mathcal U(\Delta \mathbf x) \;=\; \tfrac12\, \mathbf e_t^\top \bm{K}_t \mathbf e_t \;+\; \tfrac12\, \mathbf e_r^\top \bm{K}_r \mathbf e_r \;\in \mathbb R_{\ge 0}.
\end{equation}
The elastic wrench \(\mathbf F_{\text{elastic}} \in \mathfrak{se}^\ast(3)\) can then be derived from the gradient
\begin{equation}\label{eq:gradient}
\mathbf F_{\text{elastic}} \;=\; \frac{\partial \mathcal U}{\partial \Delta \mathbf x} = \begin{pmatrix}
    \bm{f}_{\text{elastic}} \\[2pt] \bm{m}_{\text{elastic}}
    \end{pmatrix},
\end{equation} 
with $\bm{f}_{\text{elastic}} \in \mathbb{R}^3$ and $\bm{m}_{\text{elastic}} \in \mathbb{R}^3$ being the elastic force and moment, respectively.

\subsubsection{Damping design}\label{ap:dampingDesign}
The damping design was formulated to ensure that energy dissipation scales consistently with both the commanded stiffness $\bm{K}_{\text{\{t,r}\}}$ and the task-space inertia for translations and rotations $\bm{\Lambda}_{\text{\{t,r}\}} \in \mathbb{R}^{3 \times 3}$. To account for numerical issues near singular configurations, $\bm{\Lambda}_{\text{\{t,r}\}}$ was computed using a damped least-squares method \cite{lachner_shaping_2022}.

A user-selected positive time constant $d>0$ defines the diagonal damping ratio matrix 
\begin{equation}
    \bm{D}_{\text{\{t,r}\}} = d \,\bm{I}_3 .
\end{equation}
The symmetric positive semidefinite square roots $\sqrt{\bm{\Lambda}_{\text{\{t,r}\}}}$ and $\sqrt{\bm{K}_{\text{\{t,r}\}}}$ are obtained by eigendecomposition for $\bm{\Lambda}_{\text{\{t,r}\}}$ and elementwise for $\bm{K}_{\text{\{t,r}\}}$. As presented in \cite{albu-schaffer_cartesian_2003}, an intermediate damping matrix $\bm{B}_{\text{\{t,r}\}}$ is constructed as
\begin{equation}
\begin{aligned}
\bm{B}_{\text{\{t,r\}}}
&= \sqrt{\bm{\Lambda}_{\text{\{t,r\}}}} \,\bm{D}_{\text{\{t,r\}}}\,\sqrt{\bm{K}_{\text{\{t,r\}}}} \\
&\quad + \sqrt{\bm{K}_{\text{\{t,r\}}}} \,\bm{D}_{\text{\{t,r\}}}\,\sqrt{\bm{\Lambda}_{\text{\{t,r\}}}} .
\end{aligned}
\end{equation}

From $\bm{B}_{\text{\{t,r}\}}$, a scalar damping coefficient is computed to normalize the dissipation relative to the overall stiffness:
\begin{equation}
    \lambda_{\text{\{t,r}\}} \;=\; \frac{2\,\mathrm{tr}(\bm{B}_{\text{\{t,r}\}})}{\mathrm{tr}(\bm{K}_{\text{\{t,r}\}})}
    \;=\; \frac{2\,\mathrm{tr}(\bm{B}_{\text{\{t,r}\}})}{k_x+k_y+k_z}.
\end{equation}
Finally, the damping matrices can be expressed as
\begin{equation}
    \bm{B}_t = \lambda_t \, \bm{K}_t ,
\end{equation}
for translation and 
\begin{equation}
    \bm{B}_r = \lambda_r \, \bm{K}_r ,
\end{equation}
for rotation.

\subsubsection{Control Force}
The force $\bm{f}_{\text{ctrl}} \in \mathbb{R}^3$ is composed of $\bm{f}_{\text{elastic}}$ and damping through matrix $\bm{B}_t \in \mathbb{R}^{3 \times 3}$ that acts on the end-effector translational velocity $\dot{\bm{p}} \in \mathbb{R}^3$. It can be calculated as
\begin{equation}
    \bm{f}_{\text{ctrl}} = \bm{f}_{\text{elastic}} - \bm{B}_t \ \dot{\bm{p}}.
\end{equation}  

\subsubsection{Control Moment}
 The moment $\bm{m}_{\text{ctrl}} \in \mathbb{R}^3$ is composed of $\bm{m}_{\text{elastic}}$ and damping through matrix $\bm{B}_r \in \mathbb{R}^{3 \times 3}$ that acts on the angular velocity of the end-effector $\bm{\omega} \in \mathfrak{so}(3)$. It can be calculated as
\begin{equation}
    \bm{m}_{\text{ctrl}} = \bm{m}_{\text{elastic}} - \bm{B}_r \ \bm{\omega}.
\end{equation}

\begin{figure*}[t]
    \centering
    \includegraphics[width=0.65\textwidth, trim={0 2.5cm 0 0}]{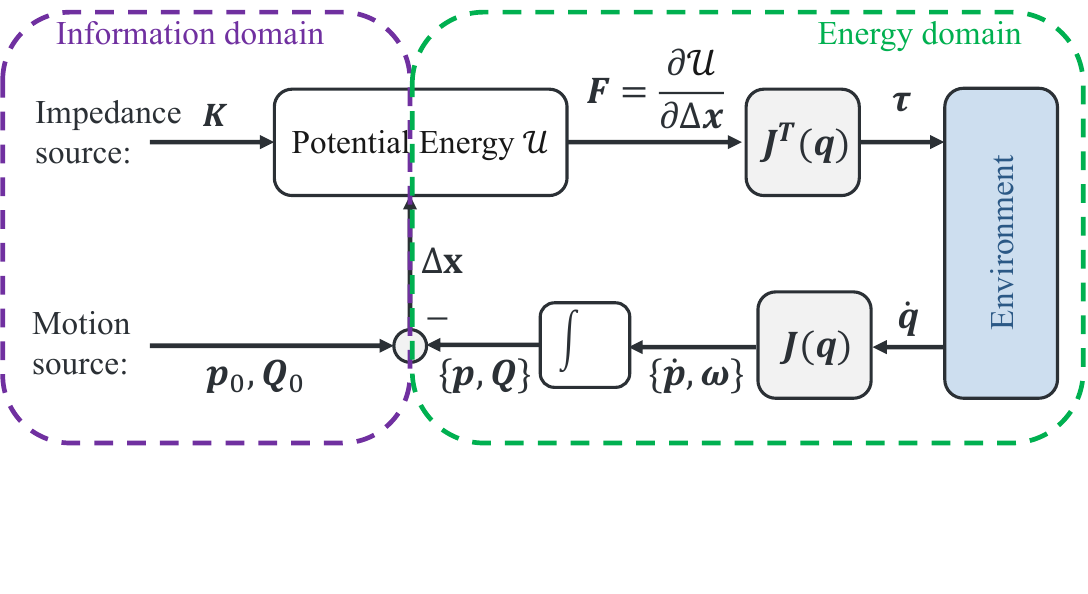}
    \caption{Norton equivalent network to represent physical interaction as the combination of a virtual motion source and an impedance source. The displacement $\Delta \bm{x}$ is coupled to stiffness $\bm{K}$ to generate wrench. }
    \label{fig:norton_equivalent}
\end{figure*}

\subsubsection{Control Torque}
Finally, the transpose of the Jacobian $\bm{J}(\bm{q}): \mathbb{R}^n \rightarrow \mathfrak{se}(3)$ provides the mapping between the 6-dimensional wrenches and the commanded joint torques $\bm{\tau}_{\text{ctrl}} \in \mathbb{R}^n$:
\begin{equation}
    \bm{\tau}_{\text{ctrl}} = \bm{J}(\bm{q})^T \begin{pmatrix}
    \bm{f}_{\text{ctrl}} \\[2pt] \bm{m}_{\text{ctrl}}
    \end{pmatrix}.
\end{equation}

In addition to the control wrench, the environment contributes external wrenches $\bm{F}_{\text{ext}} \in \mathfrak{se}^\ast(3)$. 

The  controller implemented in this paper can be found in our Github repository.

\subsubsection{Norton Equivalent Network}
The Impedance Control framework can be illustrated by the Norton Equivalent Network~\cite{hogan2014general, lachner2022geometric}, as represented in Figure~\ref{fig:norton_equivalent}. Here, two sources interact to manage physical interaction: 
\begin{itemize}
    \item The \emph{motion source}, which represents a sequence of equilibrium poses $\{\bm{p}_0, \bm{Q}_0\}$ that together form the ZFT. 
    \item The \emph{impedance source}, which represents the baseline stiffness $\bm{K}$. 
\end{itemize}
The motion source is part of the information domain. For contact-rich manipulation, it is a virtual nominal trajectory that lies beyond the physical constraints imposed by the environment. Impedance couples the motion source in the information domain to the energy domain (energy exchange through $\bm{F}$ and $\{\dot{\bm{p}}, \bm{\omega}\}$), a key property that we exploit later for energy-based stiffness estimation (Section~\ref{subsec:EnergyStiffnessEst}).

\section{Diffusion-Based Impedance Learning}

\subsection{Diffusion Model for Reconstruction of sZFT}\label{subsec:Reconstruction_sZFT}

\begin{figure}[h]
    \centering
    \includegraphics[width=0.8\columnwidth]{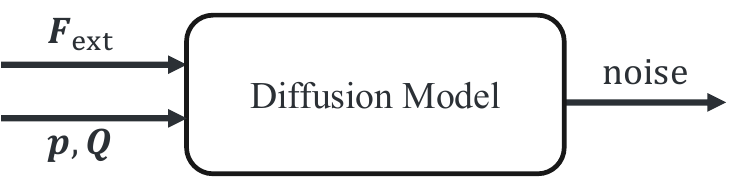}
    \caption{Diffusion-based reconstruction of the simulated Zero-Force Trajectory (sZFT). The observed pose $\{ \bm{p}, \bm{Q} \}$ together with the external wrench $\bm{F}_{\text{ext}}$ is treated as noise relative to the equilibrium $\{ \bm{p}_0, \bm{Q}_0 \}$. The Diffusion Model iteratively denoises this input to reconstruct the equilibrium trajectory, which is then used in energy-based stiffness estimation.}
    \label{fig:sZFT_diffusion}
\end{figure}

From the perspective of diffusion, the displacement $\Delta \bm{x}$ can be interpreted as noise corrupting the equilibrium pose $(\bm{p}_0,\bm{Q}_0)$. The role of the Diffusion Model is to reconstruct the equilibrium pose by iteratively removing this noise. During training, the model learns to predict the difference between the observed pose $(\bm{p},\bm{Q})$ and the ground truth equilibrium pose, guided by the measured external wrench $\bm{F}_{\text{ext}}$ (Figure~\ref{fig:sZFT_diffusion}). At inference, the model denoises $\Delta \bm{x}$ and outputs the reconstructed sZFT $(\hat{\bm{p}}_0, \hat{\bm{Q}}_0)$.

Throughout this paper, all quantities denoted by a hat $(\,\hat{\cdot}\,)$ are predictions of the Diffusion Model and correspond to the sZFT. 


It is important to distinguish between the nominal ZFT $(\bm{p}_0, \bm{Q}_0)$ and the reconstructed sZFT $(\hat{\bm{p}}_0, \hat{\bm{Q}}_0)$ in our formulation. The ZFT represents the programmed nominal motion reference and remains fixed during execution. In this work, ZFTs were generated programmatically by chaining waypoints using minimum-jerk interpolation and \emph{Spherical Linear Interpolation} (SLERP)\cite{Shoemake1985}. More generally, ZFTs can also be obtained through frame-based programming, imitation learning for free-space motion\cite{schaal1999imitation, ijspeert2013dynamical}, or teleoperation~\cite{Chi2024}. A video of the demonstrations is available on the project website and the code for ZFT generation is publicly available on GitHub.


The sZFT, in contrast, is reconstructed online by the diffusion model from measured end-effector motion and interaction wrenches. It provides a denoised estimate of the equilibrium trajectory that explains the observed interaction. During deployment, the robot continues to track the nominal ZFT as the commanded motion reference, while the reconstructed sZFT is used exclusively for impedance adaptation. Specifically, the elastic energy in the virtual springs and external work exchanged with the environment (Equations~\eqref{eq:E_t_spring} - \eqref{eq:E_t_work}) are computed with respect to the reconstructed sZFT, enabling online estimation of stiffness and damping consistent with the observed contact behavior. In this way, the diffusion model does not directly generate impedance, but instead infers the interaction-consistent equilibrium underlying the impedance controller.

\begin{figure}[h]
    \centering
    \includegraphics[width=0.7\columnwidth, trim={2.35cm 0 2.3cm 0}]{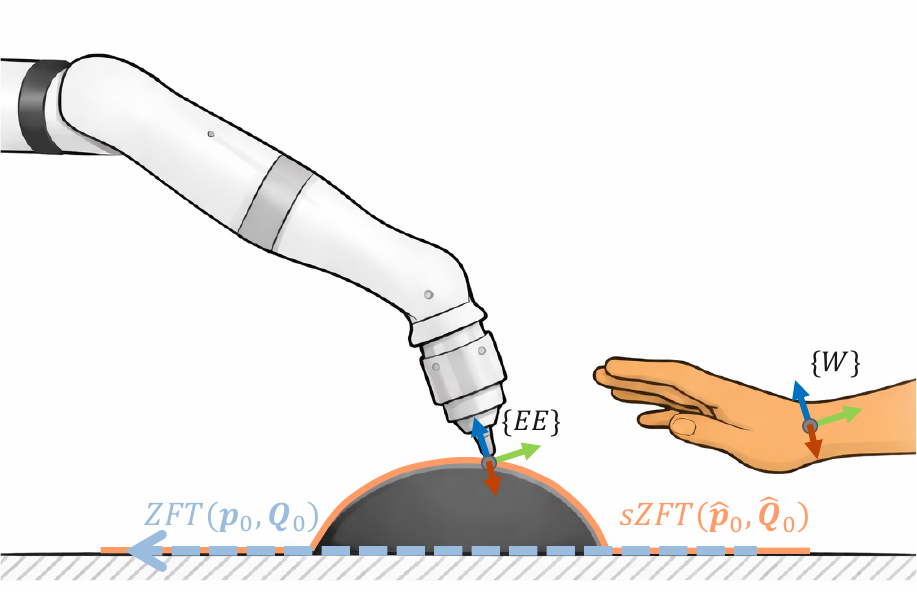}
    \caption{Illustration of the nominal Zero-Force Trajectory during deployment (ZFT, light blue) and the reconstructed simulated Zero-Force Trajectory (sZFT, skin color), generated by the Diffusion Model. The ground truth sZFT was created by teleoperation, which tracked the wrist frame \{W\} and serves as an input to the robot controller to move the end-effector frame \{EE\}. In the illustration, the ZFT is just a straight line. However, due to the external wrench during contact, the Diffusion Model will reconstruct the sZFT to imitate the demonstrated behavior. Autonomously, the robot will adapt its compliant behavior through impedance to smoothly go over the obstacle.}
    \label{fig:zft_szft}
\end{figure}

For example, when the robot encounters an unexpected obstacle, the planned ZFT drives the end-effector into contact, producing an external wrench. Conditioned on this wrench, the Diffusion Model reconstructs the sZFT and updates the controlled impedance. An illustration of the difference between ZFT and sZFT can be seen in Figure~\ref{fig:zft_szft}.

Having illustrated the effect conceptually, we now formalize the denoising process, which can be expressed in translational and rotational noise components. For translations, the full noise is defined as
\begin{equation}
    \bm{p}_{\text{noise}} = \bm{p} - \bm{p}_0 .
\end{equation}
During training, a fraction of this noise is added to the clean equilibrium pose via the noise scheduler  (Section~\ref{subsec:NoiseScheduling}), and the network is trained to predict it. At inference, the predicted noise $\hat{\bm{p}}_{\text{noise}}$ is subtracted from the observation to reconstruct the equilibrium position iteratively,
\begin{equation}
    \hat{\bm{p}}_0 = \bm{p} - \hat{\bm{p}}_{\text{noise}} .
\end{equation}

For rotations, the full noise is represented by the relative quaternion
\begin{equation}
    \bm{Q}_{\text{noise}} = \bm{Q} \, \bm{Q}_0^{-1},
\end{equation}
which encodes the rotational displacement between the observed quaternion $\bm{Q}$ and the equilibrium quaternion $\bm{Q}_0$. Partial training noise is injected to $\bm{Q}_0$ using SLERP~\cite{Shoemake1985} (Section~\ref{subsec:NoiseScheduling}), and the network is trained to predict $\hat{\bm{Q}}_{\text{noise}}$. At inference, the equilibrium quaternion is reconstructed by
\begin{equation}
    \hat{\bm{Q}}_0 = \bm{Q} \, \hat{\bm{Q}}_{\text{noise}}^{-1}.
\end{equation}

The reconstructed sZFT provides the key link between the Diffusion Model and impedance adaptation. It serves as the input for the energy-based stiffness estimation described in Section~\ref{subsec:EnergyStiffnessEst}. 

\subsection{Noise Scheduling}\label{subsec:NoiseScheduling}
For the denoising process, the maximum noise magnitude was computed separately for translations and rotations, and a variational noise schedule was applied across diffusion steps.

\subsubsection{Translations}
For translations, the deviation between the observed end-effector position $\bm{p}$ and the ground-truth ZFT $\bm{p}_0$ was treated as noise. Additional zero-mean Gaussian noise $\bm{\epsilon}_{\text{gauss}} \sim \mathcal{N}(0, \sigma^2 I)$ was injected during training to improve generalization and prevent overfitting \cite{Bishop1995}. The translational noise for the model was defined as
\begin{equation}
    \text{noise}_t = \bm{p} - \bm{p}_0 + \bm{\epsilon}_{\text{gauss}} ,
\end{equation}
with perturbed positions computed as
\begin{equation}
    \bm{p}_{\text{noise}} = \bm{p}_0 + \beta_t \cdot \text{noise}_t ,
\end{equation}
where $\beta_t$ is the noise scaling factor at diffusion step $t$.

\subsubsection{Rotations}
In our robot controller, orientations are represented with unit quaternions $\bm{Q}$. Injecting noise, as in the translational case, would violate the unit-norm constraint and distort the geometry of rotations. To address this, we introduce a SLERP-based \cite{Shoemake1985} noise scheduler that perturbs quaternions directly on the unit sphere. This approach preserves valid orientations, provides perturbations consistent with geodesics, and allows rotational noise to be scheduled smoothly across diffusion steps. To our best knowledge, this is the first physically-consistent noise scheduler for rotations in robotics, formulated directly on a global representation of rotations \cite{lachner2022geometric}. 

\begin{figure}[h]
    \centering
    \includegraphics[width=0.4\columnwidth, trim={0cm 0 0cm 0}]{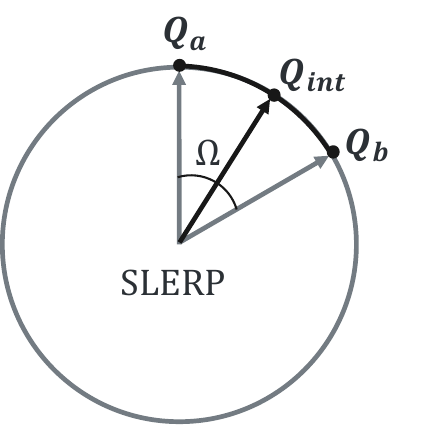}
    \caption{Rotational noise scheduling with Spherical Linear Interpolation (SLERP) \cite{Shoemake1985}. Perturbed quaternions stay on the unit sphere, enabling geometry-consistent noise injection and removal.}
    \label{fig:slerp}
\end{figure}

The SLERP process is illustrated in Figure~\ref{fig:slerp}. The interpolation between two quaternions $\bm{Q}_a$ and $\bm{Q}_b$ is defined as
\begin{equation}
\begin{aligned}
\text{SLERP}(\bm{Q}_a, \bm{Q}_b; \mu)
&= \frac{\sin((1-\mu)\Omega)}{\sin(\Omega)} \bm{Q}_a \\
&\quad + \frac{\sin(\mu \Omega)}{\sin(\Omega)} \bm{Q}_b .
\end{aligned}
\end{equation}

where $\Omega = \arccos(\langle \bm{Q}_a, \bm{Q}_b \rangle)$ is the angle between $\bm{Q}_a$ and $\bm{Q}_b$, and $\mu \in [0,1]$ is the interpolation factor. 

\begin{figure*}[t]
    \centering
    \includegraphics[width=0.95\textwidth]{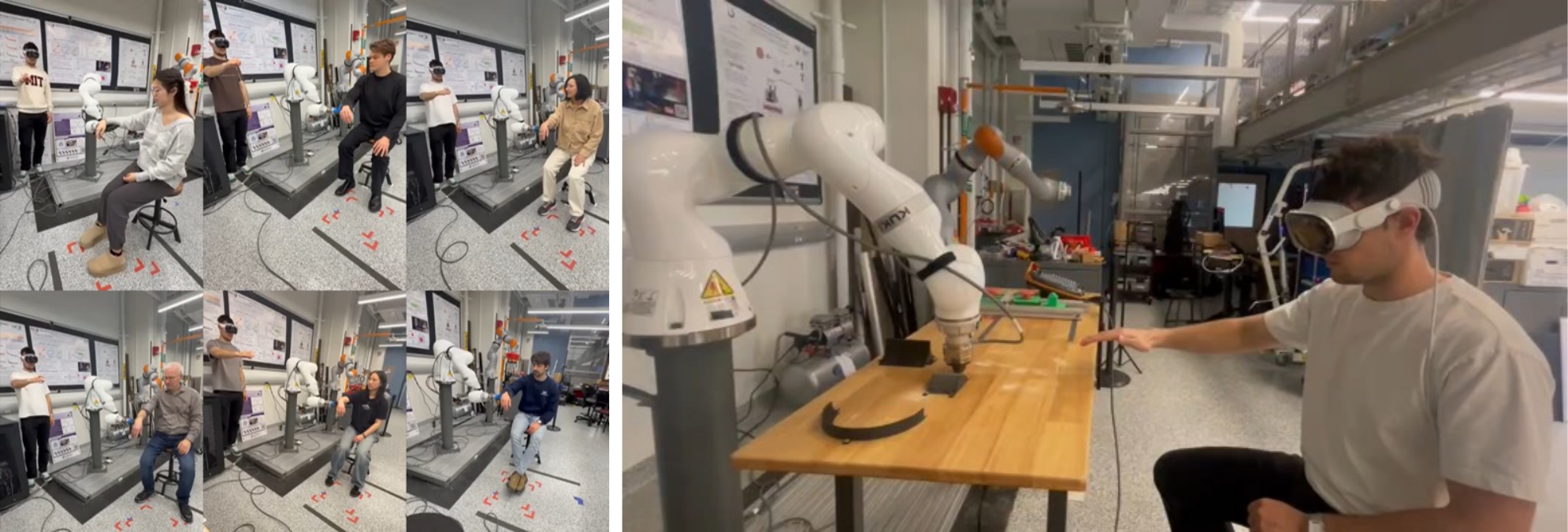}
    \caption{Examples of teleoperated demonstrations used for training. (Left) Physical therapy dataset with robotic-assisted upper-limb tasks. (Right) Parkour dataset with free-space and contact-rich motions across obstacles.}
  \label{fig:dataCollection}
\end{figure*} 

\subsection{Data Collection}\label{subsec:DataCollection}

All training data for Diffusion-Based Impedance Learning were collected through teleoperation using the VisionProTeleop framework~\cite{park2024} and Apple Vision Pro (AVP), which enabled markerless and spatially unconstrained teleoperation without external infrastructure. The AVP continuously tracked the operator’s hand motion in six degrees of freedom, allowing teleoperation in free space, during contact interactions, and throughout transitions into and out of contact (Figure~\ref{fig:dataCollection}). Within the framework, the operator’s hand pose was represented as a homogeneous transformation matrix based on a 27-joint kinematic hand model. In our implementation, the \texttt{handMiddleFingerKnuckle} (index 11) was used as the primary keypoint for translation and rotation tracking. During teleoperation, the tracked wrist pose served as the commanded ZFT and was subsequently treated as the ground-truth sZFT for model training. In parallel, external interaction wrenches at the robot end-effector were recorded using a force/torque sensor. Pose data were transferred via shared memory between the Python-based VisionProTeleop framework and the C++ robot controller. The implementation of the teleoperation interface is publicly available in our GitHub repository.

Unlike demonstrations with manipulandums and haptic interfaces, the AVP framework allowed spatially unrestricted teleoperation. In practice, the operator’s hand pose tracked by AVP could move freely in space, unconstrained by physical objects, whereas the robot end-effector was constrained by the actual contact surfaces. Therefore, the shift between the commanded ZFT and the robot pose generated an external wrench. 

Two datasets were generated. The first consisted of a parkour-style scenario with multiple obstacles (Figure~\ref{fig:dataCollection}, right). A total of 17,457 samples were recorded. These data captured diverse interaction behaviors, including free-space motion, transitions into and out of contact, and sustained contact with unknown obstacles in an unstructured environment.

The second dataset focused on robotic-assisted upper-limb rehabilitation tasks (Figure~\ref{fig:dataCollection}, left). A total of 55,838 samples were recorded. Eight healthy adults (4 female, 4 male; age 20–64 years; height 162–192 cm; weight 64–91 kg) performed three types of movements derived from standard post-stroke rehabilitation protocols: (i) passive lifting, consisting of repeated vertical up-and-down motions of the arm; (ii) out-of-shoulder movements, involving arm motions that include shoulder rotation; and (iii) an activity-of-daily-living (ADL) task, in which subjects mimicked a feeding motion by moving the hand toward the mouth.

Each therapy type was performed under three interaction conditions designed to elicit different levels of subject participation: 1) passive condition: subjects were instructed to relax their arm and allow the robot to move it without resistance; 2) cooperative condition: subjects observed the motion demonstrated by the teleoperator wearing the AVP and actively attempted to minimize the interaction forces between their arm and the robot hand; 3) mixed condition: subjects alternated between relaxing their arm and actively cooperating with the robot motion. Together, these conditions were chosen to capture varying levels of subject participation and assistance, providing diverse interaction behaviors for learning contact-consistent impedance adaptation. 

The study was approved by MIT’s Committee on the Use of Humans as Experimental Subjects (COUHES), with informed consent obtained from all participants. This dataset was used to train the model presented in this paper; however, the deployment of upper-limb rehabilitation applications is part of a separate project and is not reported here.

The desired compliant robot behavior was chosen during the data collection phase and kept unchanged during the deployment phase (Section~\ref{sec:experiments}). For all trials, the baseline translational stiffness was set to $800 \ \frac{\text{N}}{\text{m}}$, and the baseline rotational stiffness was set to $150 \ \frac{\text{Nm}}{\text{rad}}$.

\subsection{Model Selection to Reconstruct sZFT}\label{subsec:ModelSelection}
The model selection was guided by two initial requirements: (i) sufficient accuracy in reconstructing both translational and rotational components and (ii) a small model capacity, suitable for real-time deployment. To evaluate this trade-off, we varied hidden dimensions, number of layers, attention heads, and loss-weighting strategies. 

Performance was assessed using translational error (positional loss in mm) and rotational error (angular magnitude $\theta$ about unit-axis and unit-axis deviation $\alpha$ in degrees). 

\begin{figure}[h]
    \centering
    \includegraphics[width=0.6\linewidth]{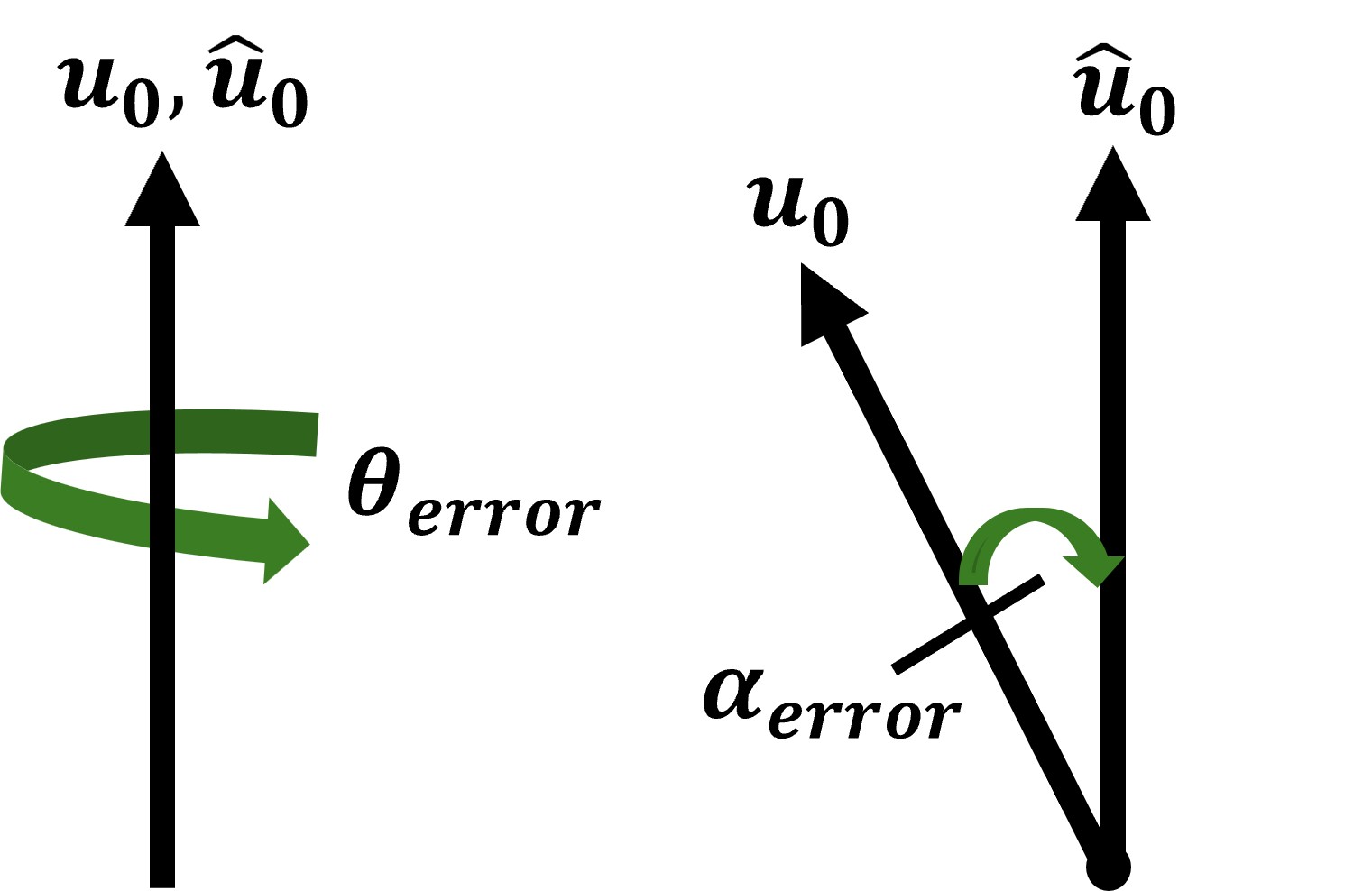}
    \caption{Graphical representation of rotational losses, showing axis deviation $\alpha_{\text{error}}$ and magnitude error $\theta_{\text{error}}$.}
    \label{fig:losses}
\end{figure}

Figure~\ref{fig:losses} provides a graphical representation of the rotational losses, illustrating the decomposition into axis deviation $\alpha_{\text{error}}$ and magnitude error $\theta_{\text{error}}$.

The initial requirements were \SI{2}{\milli\meter}  for the translational error and $2.8^\circ$ (half of the small-angle approximation) for the angular error and unit-axis deviation. These values were determined heuristically based on previous contact-rich experiments \cite{Lachner2025}.

Table~\ref{tab:transformerResults} reports the results for the Transformer-based Diffusion Model on the parkour dataset. The configuration with 512 hidden dimensions, 4 attention heads, and 6 layers achieved the best balance between accuracy and efficiency, with positional error of $0.994$ mm, $\theta$ error of $0.249^\circ$, and $\alpha$ error of $0.003^\circ$. Increasing the loss weight of the $\theta$ component (factor 5) yielded marginal improvements in rotational accuracy ($\theta = 0.240^\circ$, $\alpha = 0.003^\circ$) without degrading translational fidelity. 
\begin{table*}[t]
    \centering
    \caption{Transformer-based Diffusion Model results (parkour dataset). The selected model (highlighted) combines high accuracy with low computational complexity.}
    \begin{tabular}{cccccc}
        \toprule
        \textbf{Hidden Dim} & \textbf{Heads} & \textbf{Layers} & $\bm{\theta}$ \textbf{Loss [deg]} & $\bm{\alpha}$ \textbf{Loss [deg]} & \textbf{Positional Loss [mm]} \\
        \midrule
        512  & 4 & 4 & 0.307 & 0.004 & 1.047 \\
        \rowcolor{gray!15} 512  & 4 & 6 & 0.249 & 0.003 & 0.994 \\
        512  & 8 & 4 & 0.282 & 0.004 & 1.030 \\
        512  & 8 & 6 & 0.286 & 0.003 & 0.989 \\
        1024 & 4 & 4 & 0.228 & 0.003 & 1.033 \\
        1024 & 4 & 6 & 0.224 & 0.003 & 1.033 \\
        1024 & 8 & 4 & 0.227 & 0.003 & 1.016 \\
        1024 & 8 & 6 & 0.288 & 0.004 & 1.057 \\
        \bottomrule
    \end{tabular}
    \label{tab:transformerResults}
\end{table*}

A notable result was that retraining this model on a combined dataset of parkour and robotic-assisted therapy motions--two completely different physical interaction tasks--further improved accuracy (positional error $0.883$ mm, $\theta$ error $0.233^\circ$, $\alpha$ error $0.002^\circ$). Importantly, the two datasets were collected with different tools and inertial properties (lightweight tool for parkour and a 600 g prosthetic hand for physical therapy) yet the model compensated for these differences without explicit training. This robustness highlights the ability of Diffusion-Based Impedance Learning to generalize across different tasks and tool inertial properties on a single robot platform. The selected architecture was ultimately adopted as the final model because it satisfied the accuracy requirements while remaining small enough for real-time torque control.

Another notable finding of this work is the small amount of training data required. The model already showed strong performance on the parkour dataset with only tens of thousands of samples ($17{,}000$ samples). For the combined data set, $72{,}000$ samples at a controller rate of $0.005 \ \text{s}$ correspond to just 1.6 hours of demonstration. This is in contrast to image-based Diffusion Models, which require $100{,}000$ samples even for learning single manipulation skills such as block pushing \cite{Chi2024, Carvalho2023}.

\subsection{Energy-based Stiffness Estimation}\label{subsec:EnergyStiffnessEst}
Impedance control can be interpreted as virtual springs that pull the robot end-effector toward an equilibrium pose. When the robot interacts with the environment, the deformation of these virtual springs is transformed into external wrenches, which perform mechanical work. A shift of the reconstructed sZFT by the Diffusion Model changes the deformation, and thus the elastic energy stored in the virtual springs. By relating this change in elastic energy to the measured work, we can compute the stiffness equivalent that reproduces the observed interaction behavior.

For deployment, a user specifies baseline translational and rotational stiffness values that define the nominal interaction behavior. During execution, the Diffusion Model predicts the sZFT, which represents the equilibrium pose correction, consistent with the current interaction wrench. Rather than modifying the nominal ZFT, we adapt the impedance by reducing stiffness equivalent to the equilibrium reconstruction via sZFT.

Specifically, we compute nonnegative stiffness reduction terms $k_{t,i}^{\star}$ and $k_{r,i}^{\star}$ for each translational and rotational axis $i \in \{x,y,z\}$. These terms are subtracted from the predefined baseline stiffness values $K_{t,i,\text{max}}$ and $K_{r,i,\text{max}}$, yielding the effective stiffness used by the controller:
\begin{subequations}
\begin{align}
    k_{t,i} &= K_{t,i,\text{max}} - k_{t,i}^{\star}, \\
    k_{r,i} &= K_{r,i,\text{max}} - k_{r,i}^{\star}.
\end{align}
\end{subequations}

In our experiments, the baseline stiffness values were set deliberately high ($K_{t,i,\text{max}} = 800 \ \text{N/m}$, $K_{r,i,\text{max}} = 150 \ \text{Nm/rad}$). High baseline stiffness is desirable for tasks that require significant elastic energy storage, such as pressing or inserting a workpiece into another. We therefore selected baseline stiffness values that the robot can realistically render and that are sufficient for all tasks considered. The Diffusion Model then autonomously reduced these stiffnesses when interaction occurred, adapting the impedance only where necessary.

\subsubsection{Translations}
For translations, the displacement was defined as $\hat{\bm{e}}_t = \bm{p} - \hat{\bm{p}}_0$. The elastic energy $E_t \in \mathbb{R}$ of a virtual translational spring is
\begin{equation}\label{eq:E_t_spring}
    E_t^{\text{spring}} = \tfrac{1}{2} \bm{K}_t \lVert \hat{\bm{e}}_t \rVert^2.
\end{equation}
Here, $\lVert . \lVert$ denotes the Euclidean norm. The work done by the external force $\bm{f}_{\text{ext}} \in \mathbb{R}^3$ along the virtual displacement $\hat{\bm{e}}_t$ is given by the inner product $\langle .,.  \rangle$:
\begin{equation}\label{eq:E_t_work}
    E_t^{\text{work}} = \langle \bm{f}_{\text{ext}}, \hat{\bm{e}}_t \rangle .
\end{equation}
Equating the right hand sites of Equation~\eqref{eq:E_t_spring} with Equation~\eqref{eq:E_t_work} yields the translational stiffness contribution
\begin{equation}
    \bm{k}_t^{\star} = \frac{2 \langle \bm{f}_{\text{ext}}, \hat{\bm{e}}_t \rangle}{\lVert \hat{\bm{e}}_t \rVert^2}.
\end{equation}
To avoid oscillatory behavior, a damping term proportional to end-effector velocity was introduced:
\begin{equation}
    \tilde{\bm{e}}_t = \kappa_t \ \hat{\bm{e}}_t - \gamma_t \ \dot{\bm{p}} ,
\end{equation}
where $\kappa_t$ is a unit translational stiffness, $\dot{\bm{p}}$ is the translational end-effector velocity, and $\gamma_t$ a damping coefficient.

\subsubsection{Rotations}
For rotations, the unit axis–angle representation $\hat{\bm{e}}_r = \hat{\bm{u}}_0 \ \hat{\theta}_0$ was used (Equation~\eqref{eq:unitAxisAngle}). The corresponding elastic energy $E_r \in \mathbb{R}$ of the virtual rotational spring is
\begin{equation}\label{eq:E_r_spring}
    E_r^{\text{spring}} = \tfrac{1}{2} \bm{K}_r \lVert \hat{\bm{e}}_r \rVert^2.
\end{equation}
The work done by the external moment $\bm{m}_{\text{ext}}  \in \mathbb{R}^3$ about the rotational displacement $\hat{\bm{e}}_r$ can be calculated by
\begin{equation}\label{eq:E_r_work}
    E_r^{\text{work}} = \langle \bm{m}_{\text{ext}}, \hat{\bm{e}}_r \rangle.
\end{equation}
Equating the right hand sites of Equation~\eqref{eq:E_r_spring} with Equation~\eqref{eq:E_r_work} results in the rotational stiffness contribution
\begin{equation}
    \bm{k}_r^{\star} = \frac{2 \langle \bm{m}_{\text{ext}}, \hat{\bm{e}}_r \rangle}{\lVert \hat{\bm{e}}_r \rVert^2}.
\end{equation}
Dynamic effects were addressed by augmenting the error with a damping term:
\begin{equation}
    \tilde{\bm{e}}_r = \kappa_r \ \hat{\bm{e}}_r - \gamma_r \ \bm{\omega} ,
\end{equation}
where $\kappa_r$ is a unit rotational stiffness, $\bm{\omega}$ is the angular end-effector velocity, and $\gamma_r$ a damping coefficient.

\subsubsection{Algorithm} 
Algorithm~\ref{alg:stiffness} summarizes the procedure of the stiffness estimation. Stiffness was computed over a sliding window of 16 samples at 200 Hz (80 ms), which was sufficient given that the tasks were not highly dynamic. 
Thresholds of 1 N for force and 1 Nm for moment were applied, and values below these limits were ignored to suppress phases of low interaction.

\begin{algorithm}[h]
\caption{Energy-Based Stiffness Estimation}
\label{alg:stiffness}
\begin{algorithmic}[1]
\State \textbf{Input:} $\begin{aligned}[t]
 & \hat{\bm{e}}_t, \hat{\bm{e}}_r, \dot{\bm{e}}, \bm{\omega}, \bm{f}_{\text{ext}}, \bm{m}_{\text{ext}}, 
 \kappa_t, \kappa_r, \gamma_t, \gamma_r, \\
 & f_{\text{thres}}, m_{\text{thres}}, \bm{K}_{t,\text{max}}, \bm{K}_{r,\text{max}}, \varepsilon
\end{aligned}$
\For{$i \in \{x,y,z\}$}
    \State $\tilde{e}_{t,i} \gets \kappa_t \, \hat{e}_{t,i} - \gamma_t \, \dot{e}_i$
    \If{$|f_i| < f_{\text{thres}} \ \textbf{or} \ \tilde{e}_{t,i}^2 < \varepsilon$}
        \State $k_{t,i}^{\star} \gets 0$
    \Else
        \State $k_{t,i}^{\star} \gets \dfrac{2 \, f_i \, \tilde{e}_{t,i}}{\tilde{e}_{t,i}^2 + \varepsilon}$
        \State $k_{t,i}^{\star} \gets \max(0,\, k_{t,i}^{\star})$
    \EndIf
    \State $K_{t,i} \gets \operatorname{clip}\!\big(K_{t,\text{max},i} - k_{t,i}^{\star},\, 0,\, K_{t,\text{max},i}\big)$

    \State $\tilde{e}_{r,i} \gets \kappa_r \, \hat{e}_{r,i} - \gamma_r \, \omega_i$
    \If{$|m_i| < m_{\text{thres}} \ \textbf{or} \ \tilde{e}_{r,i}^2 < \varepsilon$}
        \State $k_{r,i}^{\star} \gets 0$
    \Else
        \State $k_{r,i}^{\star} \gets \dfrac{2 \, m_i \, \tilde{e}_{r,i}}{\tilde{e}_{r,i}^2 + \varepsilon}$
        \State $k_{r,i}^{\star} \gets \max(0,\, k_{r,i}^{\star})$
    \EndIf
    \State $K_{r,i} \gets \operatorname{clip}\!\big(K_{r,\text{max},i} - k_{r,i}^{\star},\, 0,\, K_{r,\text{max},i}\big)$
\EndFor
\State Output: $\bm{K}_t, \bm{K}_r$
\end{algorithmic}
\end{algorithm}

\subsubsection{Directional Adaptation} 
For deployment of our model in real-world experiments, the energy-based estimator was extended with a directional adaptation step. A uniform reduction of stiffness in response to external perturbations can prevent task execution, \eg the end-effector may stop when encountering an obstacle. To avoid this problem, stiffness was modulated according to the directional relevance of each axis with respect to the generated sZFT of the Diffusion Model. The alignment factors below are defined relative to $\hat{\bm{p}}_0$ and $\hat{\bm{u}}_0 \hat{\theta}_0$ contained in $\hat{\bm{e}}_t$ and $\hat{\bm{e}}_r$, which are outputs of the Diffusion Model. Hence, the adaptation mechanism depended critically on the reconstructed sZFT: if the nominal free-space ZFT were used directly, the alignment factors would not reflect the equilibrium under contact, and directional adaptation would fail (Section~\ref{sec:experiments}). 

For translations, the alignment factor
\begin{equation}
    \psi_{t,i} = \frac{|\hat{e}_{t,i}|}{\lVert \hat{\bm{e}}_t \rVert}, \quad i \in \{x,y,z\}
\end{equation}
quantifies the contribution of each axis to the intended displacement. Analogously, for rotations, we define
\begin{equation}
    \psi_{r,i} = \frac{|\hat{e}_{r,i}|}{\lVert \hat{\bm{e}}_r \rVert}, \quad i \in \{x,y,z\}.
\end{equation}

The relative translational and rotational factors $\rho_{t,i}$ and $\rho_{r,i}$ ensure that stiffness is decreased more strongly in directions that contribute little to the desired motion, while remaining close to the baseline stiffness along task-relevant axes:
\begin{subequations}\label{eq:relativeFactors}
\begin{align}
    \rho_{t,i} &= 1 - \psi_{t,i}, \\
    \rho_{r,i} &= 1 - \psi_{r,i} .
\end{align}
\end{subequations}

The final adapted stiffness values $k_{t,i}$ and $k_{r,i}$ were implemented as:
\begin{subequations}
\begin{align}
    k_{t,i} &= K_{t,i,\text{max}} - \rho_{t,i} \, k_{t,i}^{\star}, \\
    k_{r,i} &= K_{r,i,\text{max}} - \rho_{r,i} \, k_{r,i}^{\star} .
\end{align}
\end{subequations}

As will be shown in the next section, this mechanism allowed the robot to move smoothly over obstacles in the parkour task and to achieve successful in peg-in-hole insertion.

\begin{figure*}[h]
    \centering
    \includegraphics[width=0.98\textwidth, trim={0cm 0.5cm 0cm 0cm}]{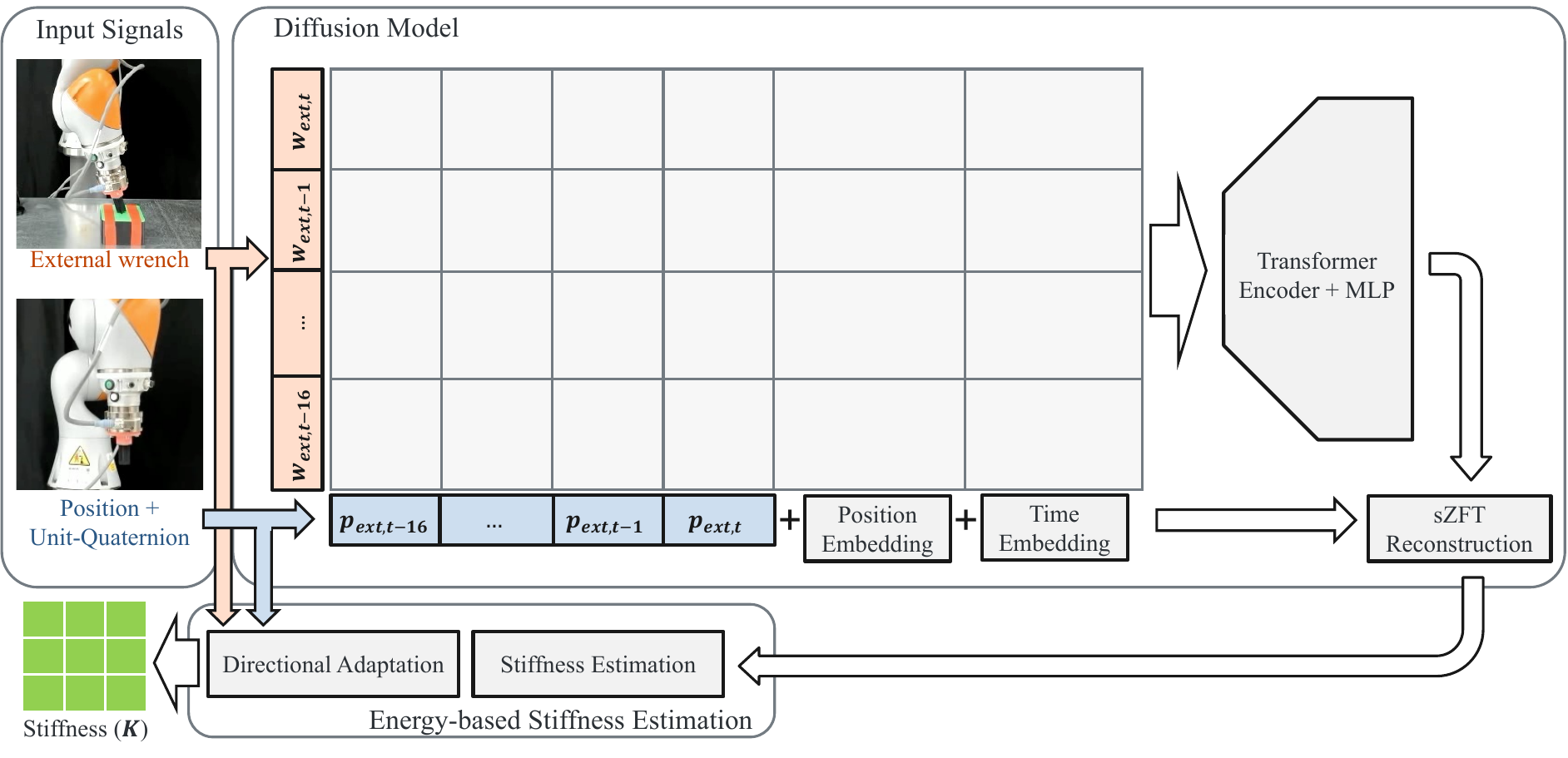}
    \caption{Overview of the proposed architecture and workflow. External wrenches and end-effector pose (position and unit-quaternion) are processed as time-series inputs to a conditional Transformer-based Diffusion Model. External wrenches are embedded as context tokens, while pose tokens are embedded with learned positional and diffusion-time encodings. The Diffusion Model reconstructs a simulated Zero-Force Trajectory (sZFT), representing a contact-consistent equilibrium. This reconstructed sZFT, together with the measured interaction signals, is passed to an energy-based stiffness estimator. Finally, directional stiffness adaptation is applied according to the directional relevance of each coordinate axis.}
    \label{fig:architecture}
\end{figure*}

An overview of the proposed model architecture and workflow can be seen in Figure~\ref{fig:architecture}.

\section{Deployment of Diffusion Model on Real Robot}\label{sec:experiments}
The model was deployed in two robotic use-cases designed to evaluate its performance in contact-rich scenarios. In our experiments, we deliberately focused on Impedance Control rather than motion generation, since many real-world applications require active compliance, \eg final assembly tasks where vision is occluded and only tactile sensing is available. The first was a parkour-style task requiring continuous interaction while traversing unknown obstacles. The second was a peg-in-hole insertion task with increasing geometric complexity. 

The experiments were conducted on a KUKA LBR iiwa with seven DOFs, using KUKA’s Fast Robot Interface (FRI) for torque control with \SI{5}{\milli\second} sample time. Built-in gravity and Coriolis/centrifugal compensation remained active throughout. Robot kinematics and dynamics were computed via the Exp[licit]-FRI interface\footnote{\url{https://GitHub.com/explicit-robotics/Explicit-FRI}} \cite{lachner2024exp}. A task-space impedance controller was implemented (Section~\ref{subsec:IC}). The external wrenches were acquired with an ATI Gamma force/torque transducer, mounted on the robot flange. The code of the robot controller is publicly available on our public GitHub repository.

In both experiments, we compared a task-space impedance controller with constant stiffness values against our proposed diffusion-based directional stiffness adaptation. For all trials, the baseline translational stiffness was identical to that used during data collection and was set to $800 \ \frac{\text{N}}{\text{m}}$, while the baseline rotational stiffness was set to $150 \ \frac{\text{Nm}}{\text{rad}}$. The same damping design was applied throughout the experiments (Section~\ref{subsec:IC}).

All workpieces were fabricated with a PRUSA i3 MK3 3D printer using PLA filament. The expected manufacturing error of the printer is on the order of \SI{0.1}{\mm}. All CAD models used in the experiments are available on our GitHub repository. Demonstration videos of both experiments are provided on the project website.

\begin{figure*}[h]
    \centering
    \includegraphics[width=0.98\textwidth, trim={0 0cm 0 0}]{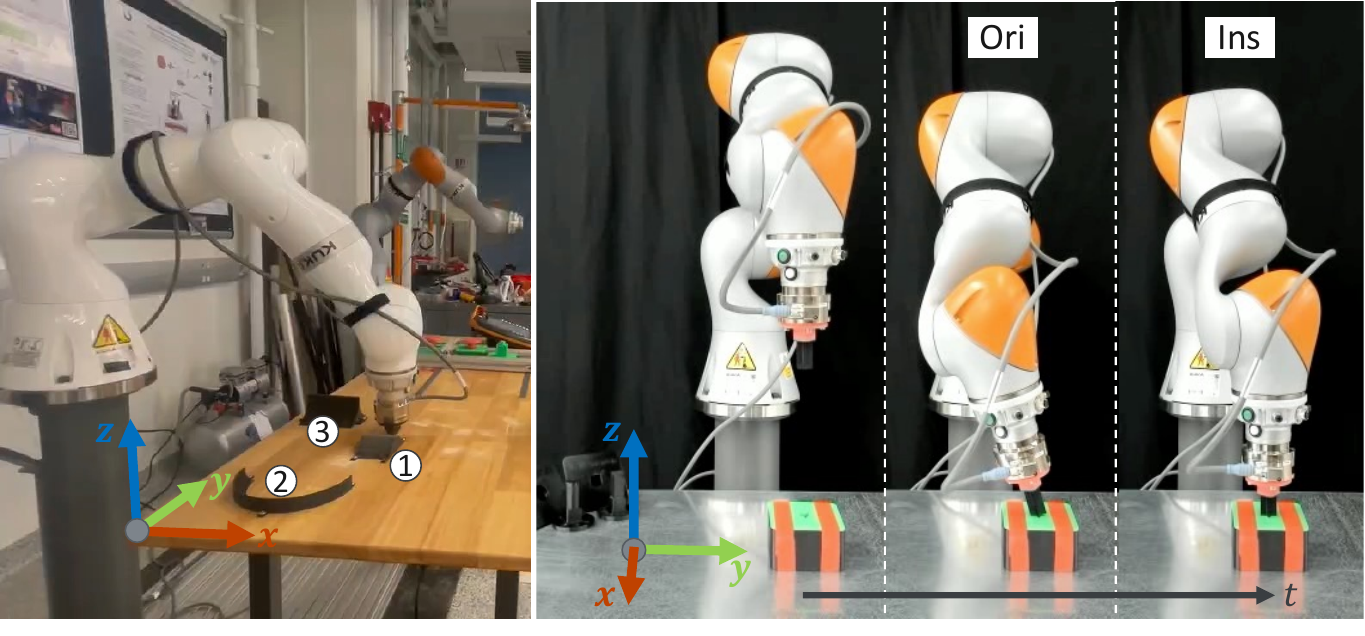}
    \caption{Phases of the two experiments and the spatially fixed coordinate frames in which they are expressed. (Left) In the parkour experiments, three obstacles are encountered (numbered 1–3). (Right) During peg insertion, the robot moved from the starting pose to inclined contact with the hole (labeled “Ori”), followed by final insertion (with the start of insertion labeled “Ins”). The colors of the major axes of the coordinate frames are shown in the experimental plots (Figure~\ref{fig:parkour} and~\ref{fig:pegInHole_star}).}
    \label{fig:frames}
\end{figure*}

All plotted entities are expressed in the spatially-fixed robot base coordinate frame. The colors of the experimental plots are consistent with the colors of the coordinate axes in Figure~\ref{fig:frames}.

\subsection{Parkour}
The first experiment consisted of a parkour-style task in which the robot traversed three obstacles while maintaining continuous contact with the table (Figure~\ref{fig:frames}, left). The desired physical interaction behavior was demonstrated by a human operator during data collection (Section~\ref{subsec:DataCollection}), who teleoperated the robot through the parkour task (Figure~\ref{fig:dataCollection}, right).

The nominal ZFT was generated by kinesthetic teaching of a number of key joint poses. A MATLAB script then computed the corresponding end-effector trajectory by smoothly interpolating between these configurations and manually shifting the trajectory along the negative z-direction to ensure constant contact with the table surface (see coordinate frame shown in Figure~\ref{fig:frames}, left). It is important to note that the resulting nominal ZFT passed through the obstacles. 


To protect the setup, stop conditions of $\lVert \bm{v} \rVert = 0.24 \ \frac{\text{m}}{\text{s}}$ and $\lVert \bm{f}_{\text{ext}} \rVert = 20$ N were defined. The constant-stiffness controller already failed at the first obstacle: the end-effector was stopped, energy accumulated in the virtual spring, and once this energy was released the end-effector accelerated into the velocity stop condition. Forces of up to 15 N were observed. In contrast, with Diffusion-Based Impedance Learning, the robot traversed all obstacles smoothly, \ie without violating the velocity or force stop conditions. The stiffness adaptation mechanism prevented jamming in directions inconsistent with the ZFT while preserving rigidity along task-relevant axes.

\begin{figure*}[h]
    \centering
    \includegraphics[width=0.95\textwidth, trim={0.5cm 0cm 0.5cm 0cm}]{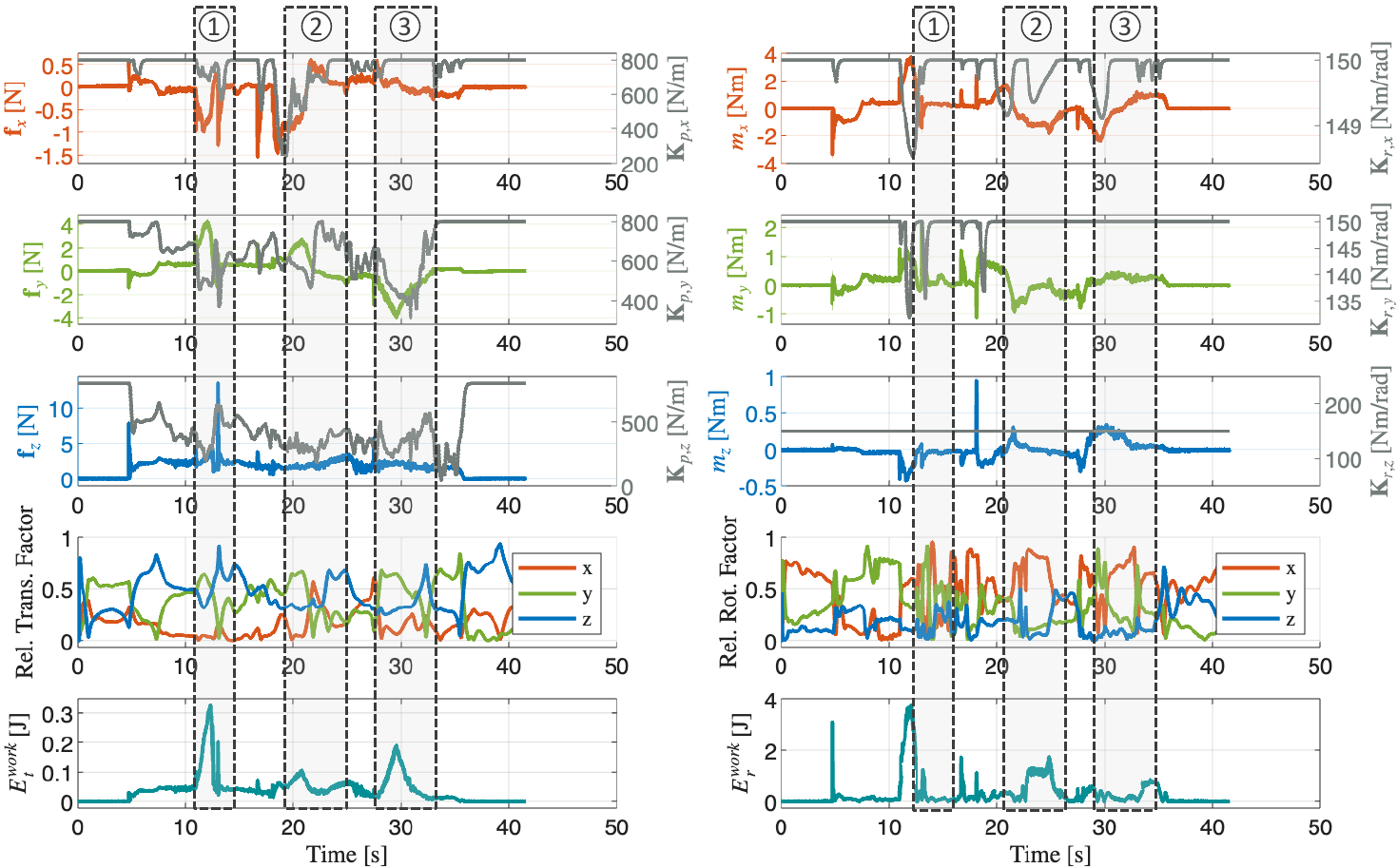}
    \caption{Measured external wrenches and commanded translational and rotational stiffness during the parkour experiment with stiffness adaptation. The relative translational and rotational factors show their respective contributions to stiffness modulation. The bottom row reports the performed translational and rotational work (Equations~\eqref{eq:E_t_work} and~\eqref{eq:E_r_work}). Shaded gray regions denote the three parkour obstacles. All quantities are expressed in the robot base frame; base-frame color coding and obstacle numbering are shown in Figure~\ref{fig:frames}.}
  \label{fig:parkour}
\end{figure*}

Figure~\ref{fig:parkour} shows the measured external wrenches together with the commanded translational and rotational stiffness values for the Diffusion-Based Impedance Learning. The bottom plots illustrate the relative translational and rotational factors (Equation~\eqref{eq:relativeFactors}) and the work done by the external force along the virtual displacement (Equations~\eqref{eq:E_t_work} and~\eqref{eq:E_r_work}).

During the entire motion, the stiffness values changed dynamically in response to external wrenches. This continuous adaptation reflected the underlying Diffusion-Based Impedance Learning, which enabled the robot to remain compliant when unexpected contact occurred while maintaining sufficient stiffness to progress through the parkour.  

Translational stiffness was adapted asymmetrically depending on the directional contribution of forces. Around 12 s, force peaks of \SI{4}{\N} along the $y$-axis and \SI{15}{\N} along the $z$-axes arose. Nevertheless, $K_{p,y}$ was reduced more strongly (reduced to near \SI{400}{\N / \m}) than $K_{p,z}$ (remaining near \SI{600}{\N / \m}) since the relative translational factor along the z-axis was high. A similar effect is visible near \SI{30}{\second}: while $f_y$ reached an absolute of nearly \SI{4}{\N} with stiffness near \SI{420}{\N / \m}, $f_z$ was only \SI{2}{\N} yet stiffness dropped to about \SI{200}{\N / \m}. These findings confirm that stiffness reduction was not simply proportional to force magnitude but was determined by the alignment with the sZFT, predicted by the Diffusion Model.  

Rotational stiffness also adapted directionally, with reductions depending on which axis contributed less to the sZFT. Around \SI{12}{\second}, moments appeared simultaneously about the $x$- and $y$-axes. Despite higher magnitudes about the $x$-axis, the stiffness $K_{r,y}$ was reduced more strongly, showing that $y$-axis rotations were less important for the intended trajectory. This selective modulation allowed the robot to maintain stability while avoiding unnecessary resistance to contact.  

\subsection{Peg-In-Hole Insertion}
The second experiment was a peg-in-hole insertion task with three different peg types: cylindrical, square, and star pegs. The nominal clearance between each peg and its corresponding hole was \SI{0.14}{\milli\meter}  for the square peg and \SI{0.20}{\milli\meter} for the circular and star pegs. All workpieces were fabricated with the same printer setup as for the parkour experiment. For all trials, task success was defined as completely inserting the peg.

For each peg type, the ZFT was generated from three manually demonstrated robot joint configurations, via kinesthetic teaching: a starting configuration, an initial contact (“touch”) configuration, and a final insertion configuration. These configurations were recorded by physically guiding the robot through the task. A smooth nominal ZFT was then generated using a MATLAB script by computing the corresponding end-effector poses and connecting them with a minimum-jerk interpolation. To induce a controlled pulling force during the final insertion phase, the equilibrium pose for final insertion was deliberately shifted along the negative $z$-direction. After completing the insertion, the robot moved straight upward along the z-axis (see coordinate frame in Figure~\ref{fig:frames}, right). Importantly, during the touch pose the peg's orientation was not aligned with the edges of the square and star pegs. As a result, the nominal ZFT did not encode any geometric alignment with the peg shape.


The three peg types represent increasing geometric complexity. The cylindrical peg is rotation-symmetric, which makes the final orientation about the end-effector $z$-axis irrelevant. In contrast, the square and star pegs require precise rotational alignment with an increasing number of edges, which significantly increases task complexity. 

This trend is reflected in the constant-stiffness experiments. The cylindrical peg succeeded in 30/30 trials. The square peg succeeded in 4/30 trials. The star peg failed in all 30 trials.

In practice, more complicated pegs like the square and star peg often require extensive manual tuning of the ZFT, of the impedance parameters, or both. This tuning is not only time-consuming but there is also no standard method that guarantees success. These results show that more complicated assembly processes require more advanced strategies.

In contrast, Diffusion-Based Impedance Learning achieved 30/30 successful insertions for all three peg types. This result was achieved by only demonstrating and interpolating three key poses per trajectory. Hence, no advanced method for trajectory generation was needed. Moreover, the outcome is particularly notable because the training data included only parkour and upper-limb rehabilitation data (Section~\ref{subsec:DataCollection}). No peg-in-hole demonstrations were used for training. Importantly, neither the Diffusion Model nor the impedance controller is provided with any information about the peg or hole geometry, pose, insertion axis, or clearance; all impedance adaptation was driven solely by measured interaction wrenches during the motion. 

Across all three peg geometries, execution times remained comparable despite the different contact conditions and insertion complexities. Mean completion times for 30 trials were $10.8 \pm 0.6$ s for the circular peg, $10.2 \pm 1.9$ s for the square peg, and $12.3 \pm 1.9$ s for the star-shaped peg.

\begin{figure*}[h]
    \centering
    \includegraphics[width=0.95\textwidth]{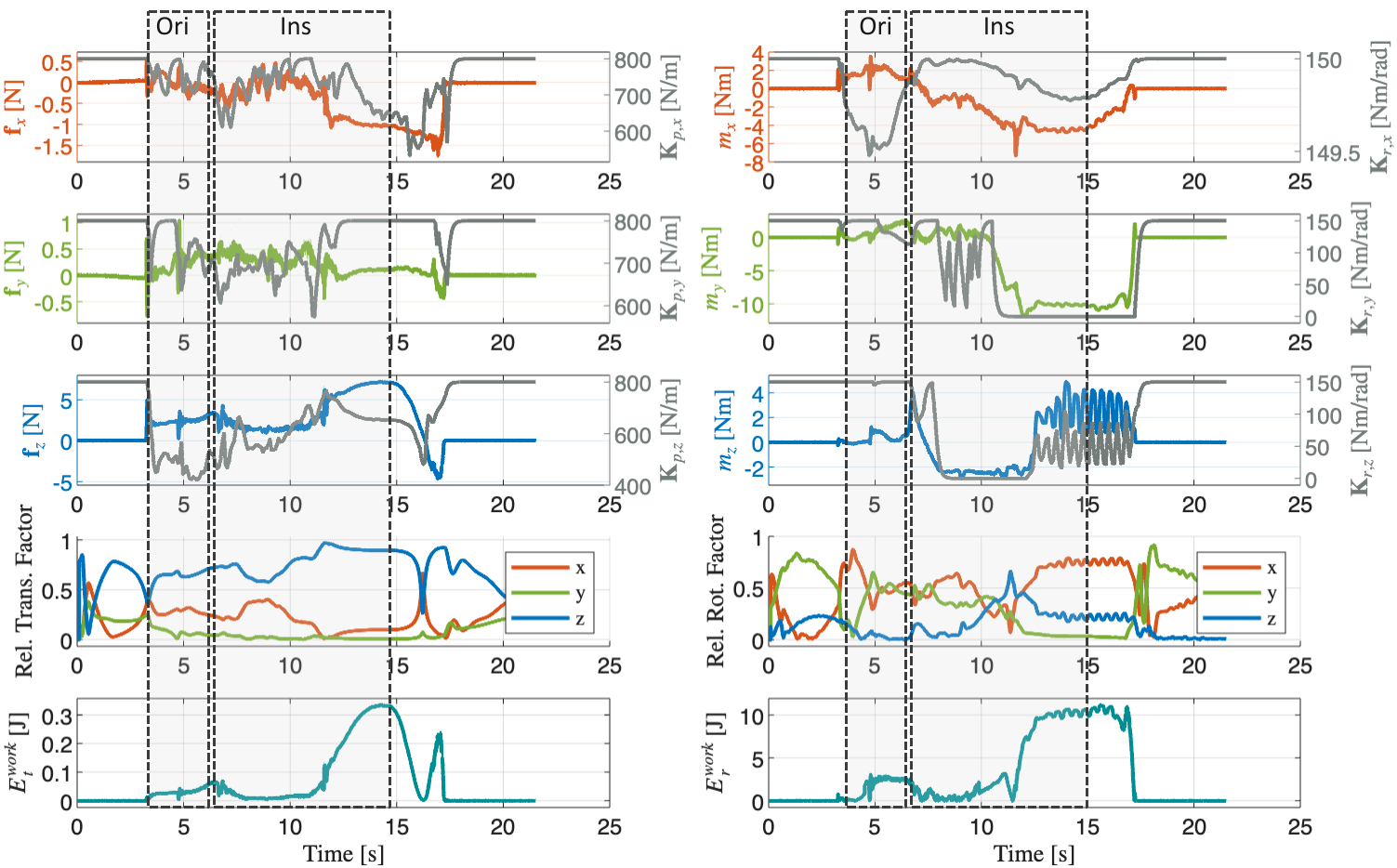}
    \caption{Measured external wrenches and commanded translational and rotational stiffness during the peg-in-hole star insertion experiment with stiffness adaptation. The relative translational and rotational factors show their respective contributions to stiffness modulation. The bottom row reports the performed translational and rotational work (Equations~\eqref{eq:E_t_work} and~\eqref{eq:E_r_work}). The shaded gray region denotes the insertion phase. All quantities are expressed in the robot base frame; base-frame color coding is shown in Figure~\ref{fig:frames}.}
  \label{fig:pegInHole_star}
\end{figure*}

Figure~\ref{fig:pegInHole_star} shows the stiffness adaptation for the star peg. Results for the other two pegs are provided in Appendix~\ref{ap:stiffnessAdaptation_pegInHole}. During all experiments, force and moment thresholds were set to $1$ N and $1$ Nm, respectively.

As illustrated in Figure~\ref{fig:pegInHole_star}, stiffness values were reduced in response to external wrenches, with the amount of reduction governed by the relative translational and rotational factors. For most of the experiment, the relative translation factor along the $z$ axis remained higher than along the $y$ axis. For instance, at around \SI{11}{\second}, small forces of \SI{0.5}{\newton} appeared along the $y$ axis. Nevertheless, the stiffness $K_{p,y}$ was reduced to \SI{600}{\N / \m}. During retraction of the peg at around \SI{16}{\second}, it can be seen that the relative translation factor along the $z$ axis dropped and consequently the stiffness $K_{p,z}$ was reduced to \SI{500}{\N / \m}. In contrast, the relative translation factor along the $x$ axis increased and consequently the stiffness $K_{p,x}$ was increased to \SI{700}{\N / \m}.

Due to the imperfect commanded ZFT, large moments about the $x$- and $y$-axes occurred during execution. The relative rotational factor about the $z$-axis remained low, leading to substantial reductions in $K_{r,z}$, which even reached \SI{0}{\N\m} between \SIrange{8}{12}{\second}. For the $x$-axis, the moments were initially small. When the moments increased significantly between \SIrange{11}{16}{\second}, the controller maintained high stiffness, since the relative rotational factor about the $x$-axis increased. A different behavior can be observed for the $y$-axis: during \SIrange{11}{17}{\second}, high moments reaching up to \SI{-10}{\N\m} occurred. Since the relative rotational factor remained low, the stiffness dropped and reached \SI{0}{\N\m}.

\subsection{Ablation Study}
In addition to the two main controllers, we performed an ablation study to isolate the contributions of the reconstructed sZFT (Section~\ref{subsec:Reconstruction_sZFT}) and the directional impedance adaptation (Section~\ref{subsec:EnergyStiffnessEst}). 

Importantly, the nominal ZFT was kept fixed throughout the ablation study. It was generated from simple kinesthetic demonstrations and was not optimized for the two tasks and pegs. Videos illustrating the kinesthetic teaching procedure for the parkour and peg-in-hole tasks are provided on the project website.

Specifically, we evaluated the following variants using the baseline impedance parameters ($800 \ \frac{\text{N}}{\text{m}}$ for translation and $150 \ \frac{\text{Nm}}{\text{rad}}$ for rotation):

\begin{enumerate}
    \item Uniform stiffness adaptation without directional adaptation, using the reconstructed sZFT
    \item Directional stiffness adaptation computed from the nominal ZFT
\end{enumerate}

Uniform stiffness adaptation without directional modulation failed in both experimental scenarios. In the parkour task, the robot was blocked at the first obstacles, while none of the peg-in-hole insertions succeeded. Directional adaptation derived from the nominal ZFT also failed in contact-rich settings. In the parkour task, the robot exceeded safety limits and stopped. In the peg-in-hole experiments, this variant achieved $30/30$ success only for the cylindrical peg, but failed for both the square and star pegs.

These results confirm that directional adaptation alone is insufficient if it is derived from the nominal ZFT. Instead, directional factors must be computed relative to the reconstructed sZFT for impedance adaptation to succeed.

To further evaluate how different uniform and non-uniform \emph{fixed} impedance parameters influence task success in the peg-in-hole experiment, we additionally evaluated the following variants:
\begin{enumerate}[label=\alph*)]
    \item Uniform high impedance: $800 \ \frac{\text{N}}{\text{m}}$ (translation), $150 \ \frac{\text{Nm}}{\text{rad}}$ (rotation)
    \item Uniform medium impedance: $400 \ \frac{\text{N}}{\text{m}}$ (translation), $50 \ \frac{\text{Nm}}{\text{rad}}$ (rotation)
    \item Uniform low impedance: $200 \ \frac{\text{N}}{\text{m}}$ (translation), $7 \ \frac{\text{Nm}}{\text{rad}}$ (rotation)
    \item Best-found, manually tuned non-uniform impedance: translation along $z$: $800 \ \frac{\text{N}}{\text{m}}$; translation along $x$ and $y$: $350 \ \frac{\text{N}}{\text{m}}$; rotation about $z$: $5 \ \frac{\text{Nm}}{\text{rad}}$; rotation about $x$ and $y$: $100 \ \frac{\text{Nm}}{\text{rad}}$
\end{enumerate}

For a) (uniform high impedance), the cylindrical peg succeeded in $30/30$ trials, the square peg in $4/30$ trials, and the star peg failed in all trials. For b) (uniform medium impedance), yielded $30/30$ success for the cylindrical peg, $13/30$ success for the square peg, and $5/30$ success for the star peg. For c) (uniform low impedance), it failed for all peg types, as the wrench required to orient the peg with the hole was insufficient.

For d) (manually tuned non-uniform impedance), the cylindrical peg succeeded in $30/30$ trails, the square peg in $26/30$ trials, and the star peg in $21/30$ trials. These parameters were identified by an experienced impedance control programmer after testing 15 different parameter combinations. In contrast, inexperienced users would likely struggle to identify a single parameter set that performs robustly across all peg geometries.

Videos of all ablation experiments for the star peg example are provided on our project website. 

A summary of the ablation study for the peg-in-hole experiment can be found in Table~\ref{tab:ablation_pih}.

\begin{table*}[t]
\centering
\caption{Peg-in-hole success rates for the ablation study. All results are reported over 30 trials.}
\label{tab:ablation_pih}
\begin{tabular}{lccc}
\toprule
\textbf{Controller Variant} 
& \textbf{Cylindrical Peg} 
& \textbf{Square Peg} 
& \textbf{Star Peg} \\
\midrule
\multicolumn{4}{l}{\emph{Time-varying Impedance}} \\
\rowcolor{gray!15} Diffusion Model (with directional adaptation) 
& 100\% & 100\% & 100\% \\
Diffusion Model w/o directional adaptation 
& 0\% & 0\% & 0\% \\
Directional adaptation w/o DM (nominal ZFT) 
& 100\% & 0\% & 0\% \\
\midrule
\multicolumn{4}{l}{\emph{Fixed Impedance}} \\
Uniform high (baseline) 
& 100\% & 13\% & 0\% \\
Uniform medium 
& 100\% & 43\% & 17\% \\
Uniform low 
& 0\% & 0\% & 0\% \\
Best-found non-uniform 
& 100\% & 87\% & 70\% \\
\bottomrule
\end{tabular}
\end{table*}

\section{Discussion}
The experimental results confirm the central goal of this work: enabling robust contact-rich manipulation by bridging the information and the energy domains and therefore unifying state-of-the-art generative AI algorithms with model-based control. In particular, the success of peg-in-hole insertions with unseen tool and task geometries demonstrates that sZFT reconstruction generalizes beyond the training distribution.

Compared to prior learning-based approaches, our framework explicitly incorporates energy transformation, enabling stable impedance adaptation under contact. Compared to traditional model-based impedance controllers, it gains adaptability by reconstructing equilibrium trajectories directly from data. This ability to bridge domains accounts for the strong performance observed in unstructured environments.

Task success in contact-rich assembly is determined by the interplay between the nominal motion reference and the generated impedance, rather than by either component alone. For the peg-in-hole experiments of this paper, the initial contact (``touch'') phase is particularly important because it establishes the conditions for the subsequent orientation and insertion phases. Nevertheless, the proposed framework exhibited considerable robustness to variations in the nominal trajectory. For both the cylindrical and square pegs, successful insertion was achieved for lateral offsets of up to approximately $\pm 3$ mm in the x and y directions. Even for the more demanding star-shaped peg, which requires accurate rotational alignment, successful insertion remained possible for offsets of approximately $\pm 1$ mm. This observation is consistent with the findings reported in~\cite{Lachner_2023_EDA}.

A key element of the framework is directional stiffness adaptation. This mechanism is generic and can be tailored to applications in other domains. One example is robotic-assisted upper-limb therapy, where stiffness must remain high along non-movement directions, while stiffness along the intended movement direction should adapt depending on whether the patient can execute the motion independently. In such settings, the same framework can be used by solely adapting the directional stiffness adaptation.

The force and moment thresholds in the stiffness estimator affect the responsiveness of impedance adaptation. In our experiments, values of 1 \si{N} and 1 \si{Nm} were selected to match the noise characteristics of the ATI Gamma force/torque sensor. These thresholds are not task-specific and can be adjusted for different sensors or applications.

At present, the adaptive impedance controller outputs diagonal, positive semidefinite stiffness matrices. This design simplifies analysis and implementation but omits cross-coupling effects that have proven beneficial in traditional solutions such as remote-center compliance (RCC) devices \cite{Whitney_1982, Hirai1990remote}. For more complex assembly tasks, integrating coupled stiffness parameters may further enhance robustness and alignment accuracy. Exploring this extension is an important direction for future work.

In this study, the baseline impedance behavior used during teleoperation was preserved during deployment, ensuring consistency between the demonstrations used for training and the interaction conditions encountered during execution. Future work should investigate robustness to changes in baseline impedance, including deployment with substantially different stiffness settings than those used during data collection. Such experiments would provide further insight into the ability of the learned model to generalize across different compliant behaviors and controller parameterizations.

The proposed approach guarantees passivity when stiffness is reduced: lowering stiffness dissipates stored potential energy and ensures that interaction with a passive environment remains output strictly passive. However, when stiffness is increased back toward the baseline, passivity is not generally guaranteed, as such changes can inject additional energy whenever displacements are nonzero. In these cases, the controller can be combined with previous work on passive impedance shaping \cite{lachner_shaping_2022} to enforce strict passivity.

In our work, we adopt a Transformer-based Diffusion Model due to its ability to capture long range temporal dependencies and flexibly condition on multimodal interaction signals. Future work should investigate a systematic comparison with alternative architectures such as convolutional or hybrid models under matched parameter and computational budgets.

Overall, our findings demonstrate that Diffusion-Based Impedance Learning is highly effective in the experimental setups investigated. While this provides strong evidence for the viability of the approach, further studies across a wider range of tasks and environments are needed to fully establish its generality. Nevertheless, the framework offers a promising design principle for stable, safe, and adaptable robot interaction in other domains including industrial assembly, physical Human-Robot Interaction, and robotic-assisted rehabilitation.

\section{Conclusion}
This work introduced \emph{Diffusion-Based Impedance Learning}, a framework that unifies generative trajectory models in the information domain with adaptive impedance regulation in the energy domain. By reconstructing simulated Zero-Force Trajectories (sZFTs) from contact-perturbed displacements and external wrench signals, the method enables real-time impedance adaptation for contact-rich manipulation.

The framework consistently outperformed fixed and heuristic strategies in experiments on a KUKA LBR iiwa. It enabled smooth traversal of obstacle-rich parkour environments where conventional impedance control failed, and achieved 30/30 success rates in cylindrical, square, and star peg-in-hole insertions without any task-specific demonstrations in the training data. The experiments in this work intentionally withhold geometric information from the controller, targeting scenarios in which such knowledge is unavailable or unreliable due to unstructured environments. Diffusion-Based Impedance Learning should therefore be understood as a general, interaction-driven alternative rather than a replacement for highly engineered solutions in fully structured assembly lines.

Beyond these results, Diffusion-Based Impedance Learning establishes a general interface between learning and model-based control. It extends the role of generative models from visuomotor trajectory planning to physical interaction, demonstrating that contact-rich manipulation can be learned from relatively small datasets while maintaining real-time control performance.

Looking ahead, we aim to extend this framework from single-arm to dual-arm setups. While this work focuses on impedance learning and used a fixed ZFT, Diffusion-Based Impedance Learning can naturally be embedded in hierarchical architectures that couple high-level visuomotor motion policies with low-level impedance regulation.


\bibliographystyle{IEEEtran}
\bibliography{bibliography}

\begin{appendices}

\section{Peg-In-Hole experiments}\label{ap:stiffnessAdaptation_pegInHole}
This appendix complements the main results by reporting stiffness adaptation 
for the circular and square pegs. The corresponding results for the star peg 
are already presented in the main text (Figure~\ref{fig:pegInHole_star}). The experimental setup, controllers, and parameters matched the main text: baseline stiffness $K_{t,i,\text{max}}=\SI{800}{N/m}$ and $K_{r,i,\text{max}}=\SI{150}{Nm/rad}$, damping as in Appendix~\ref{ap:dampingDesign}, and force/moment thresholds of \SI{1}{N} and \SI{1}{Nm}. All quantities are expressed in the robot base frame. 

For the circular peg, rotational symmetry about the end-effector $z$-axis makes the final yaw alignment irrelevant. Stiffness modulation is therefore dominated by translational factors, with only minor rotational adjustments. For the square peg, alignment with the edges increased sensitivity to off-axis contact. This yielded stronger directional reductions in both translation and rotation when axes contributed little to the sZFT, while stiffness remained high along task-relevant directions.

\begin{figure*}[h]
    \centering
    \includegraphics[width=0.9\textwidth]{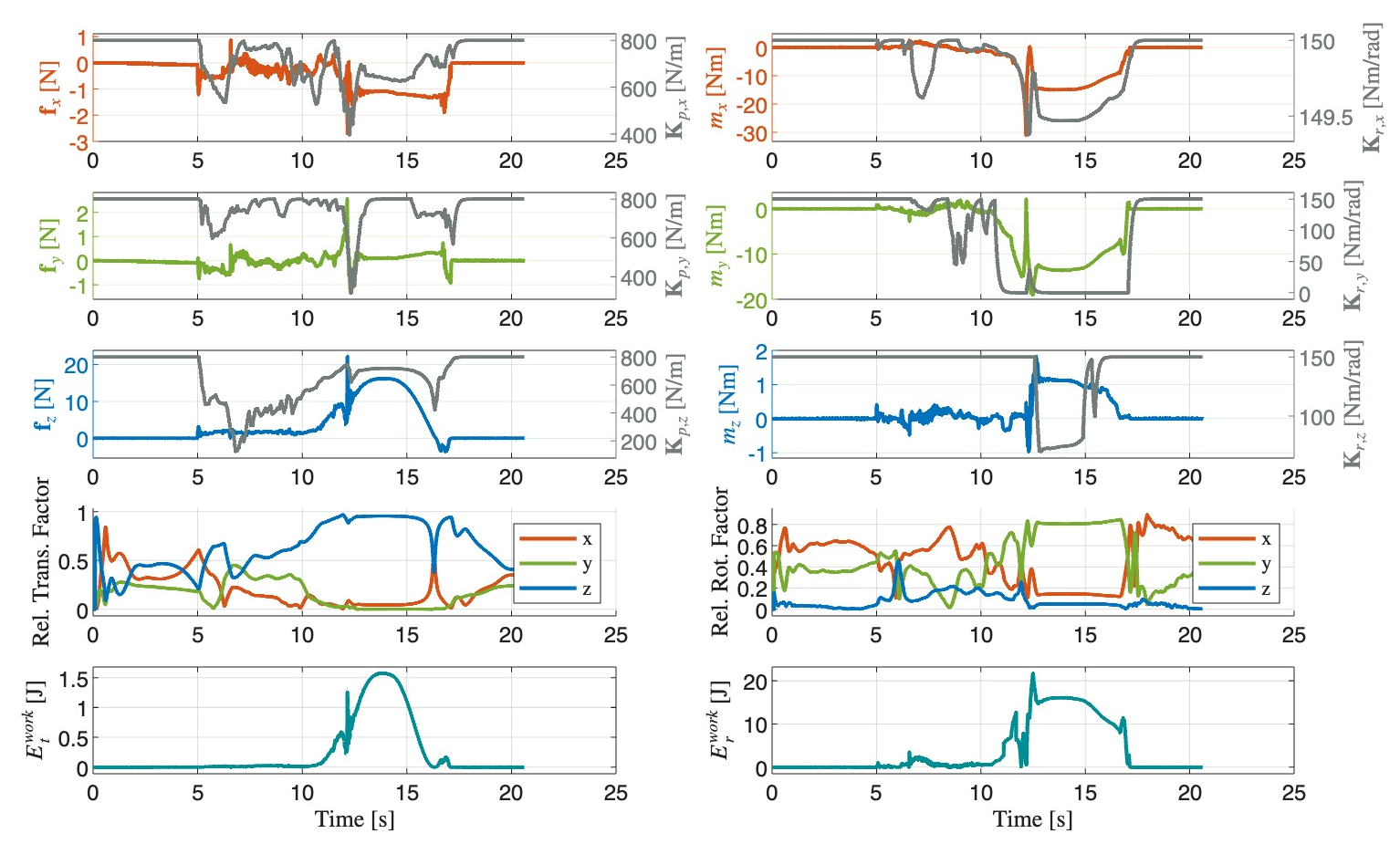}
    \caption{Cylindrical peg insertion with stiffness adaptation. External wrenches, commanded translational and rotational stiffness, and relative motion factors are shown in the robot base frame. Base-frame color coding is given in Figure~\ref{fig:frames}.}
    \label{fig:pegInHole_circ}
\end{figure*}

\begin{figure*}[h]
    \centering
    \includegraphics[width=0.9\textwidth]{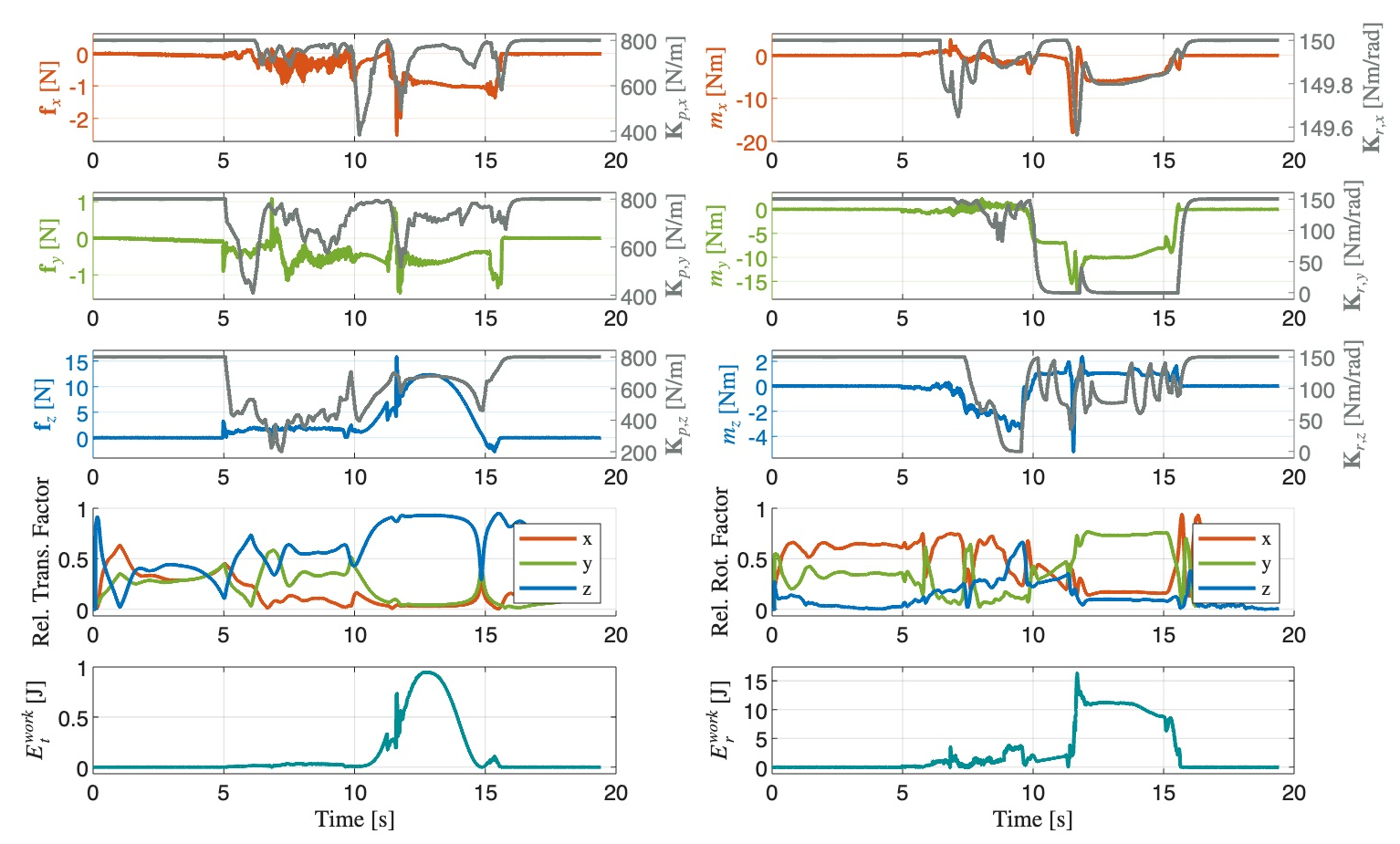}
    \caption{Square peg insertion with stiffness adaptation. External wrenches, commanded translational and rotational stiffness, and relative motion factors are shown in the robot base frame. Base-frame color coding is given in Figure~\ref{fig:frames}.}
    \label{fig:pegInHole_square}
\end{figure*}

\end{appendices}

\begin{IEEEbiography}[{\includegraphics[trim=35cm 0cm 1cm 0cm, clip, height=1.1in, keepaspectratio]{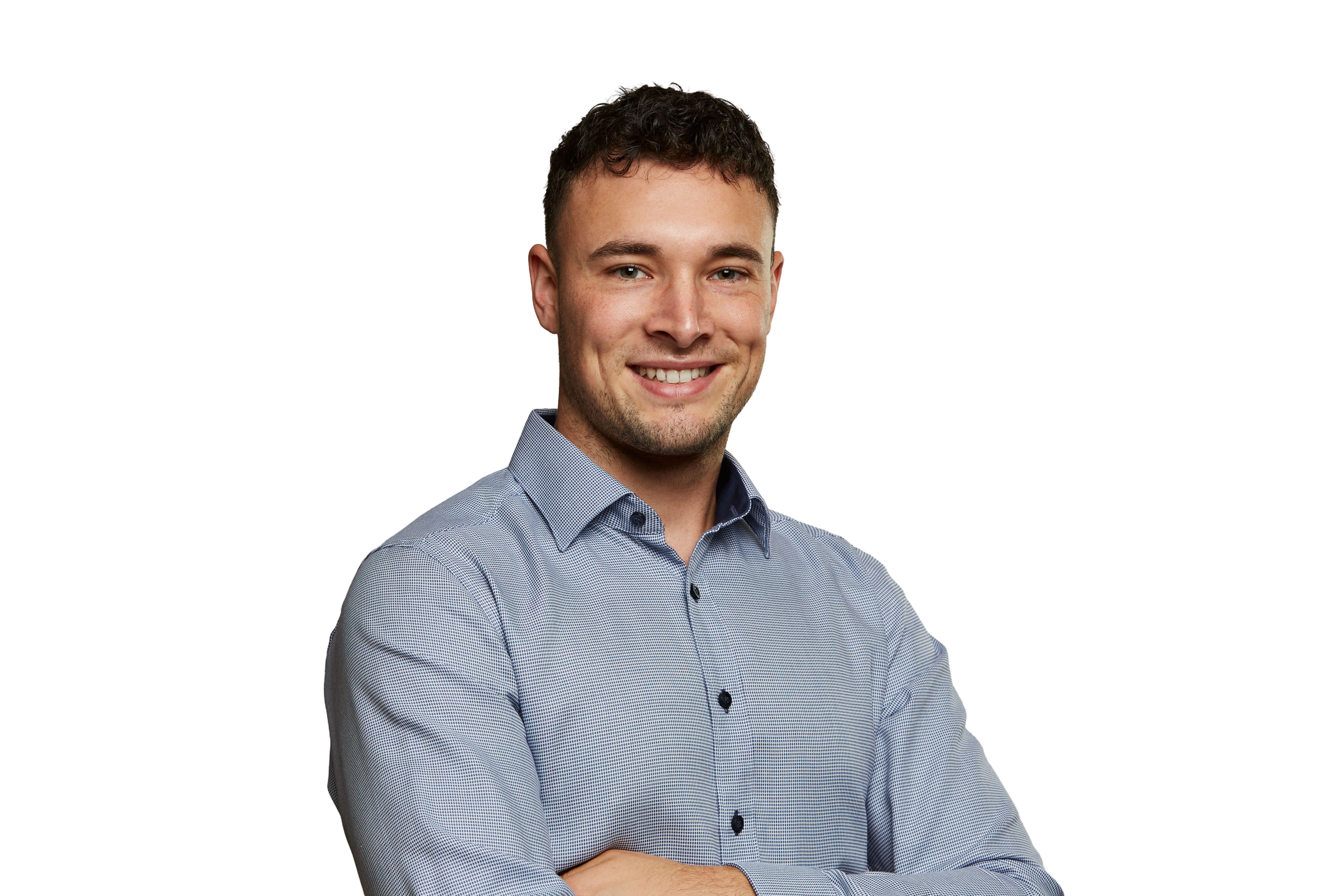}}]{Noah Geiger}
received the B.Sc. degree in Mechanical Engineering and the M.Sc. degree in Mechatronics and Information Technology, with a specialization in robotics, from the Karlsruhe Institute of Technology (KIT), Karlsruhe, Germany, in 2022 and 2025, respectively. He was a Visiting Researcher at the Massachusetts Institute of Technology (MIT), Cambridge, MA, USA. He was awarded an Engineering Scholarship from ERGO Direkt AG and an IT Scholarship from Festo GmbH \& Co. KG during his studies, as well as the DPG Abiturpreis. He began his professional experience as a Student Assistant at KIT, working on human pose estimation, and as a Working Student and Bachelor’s Candidate with Schaeffler, Karlsruhe, Germany, where he worked on machine learning for vehicle control systems. He then worked as a Data Scientist with Mercedes-Benz, Sindelfingen, Germany, focusing on scene understanding and end-to-end neural networks for autonomous driving, and as a Robotics Engineer with Bosch, Plymouth, MI, USA, where he implemented a mobile robot system for radar sensor validation. Since October 2025, he has been with the Junior Managers Program (AI/IT Track), Robert Bosch GmbH, Stuttgart, Germany, a program that prepares participants for leadership responsibility. His research interests include robotics, agentic AI, diffusion models, and physical AI. Mr. Geiger is Founder and Lead of the Studopolis Think Tank and was previously a Member of the Board of MIT VISTA.
\end{IEEEbiography}

\begin{IEEEbiography}[{\includegraphics[width=1in,clip,keepaspectratio]{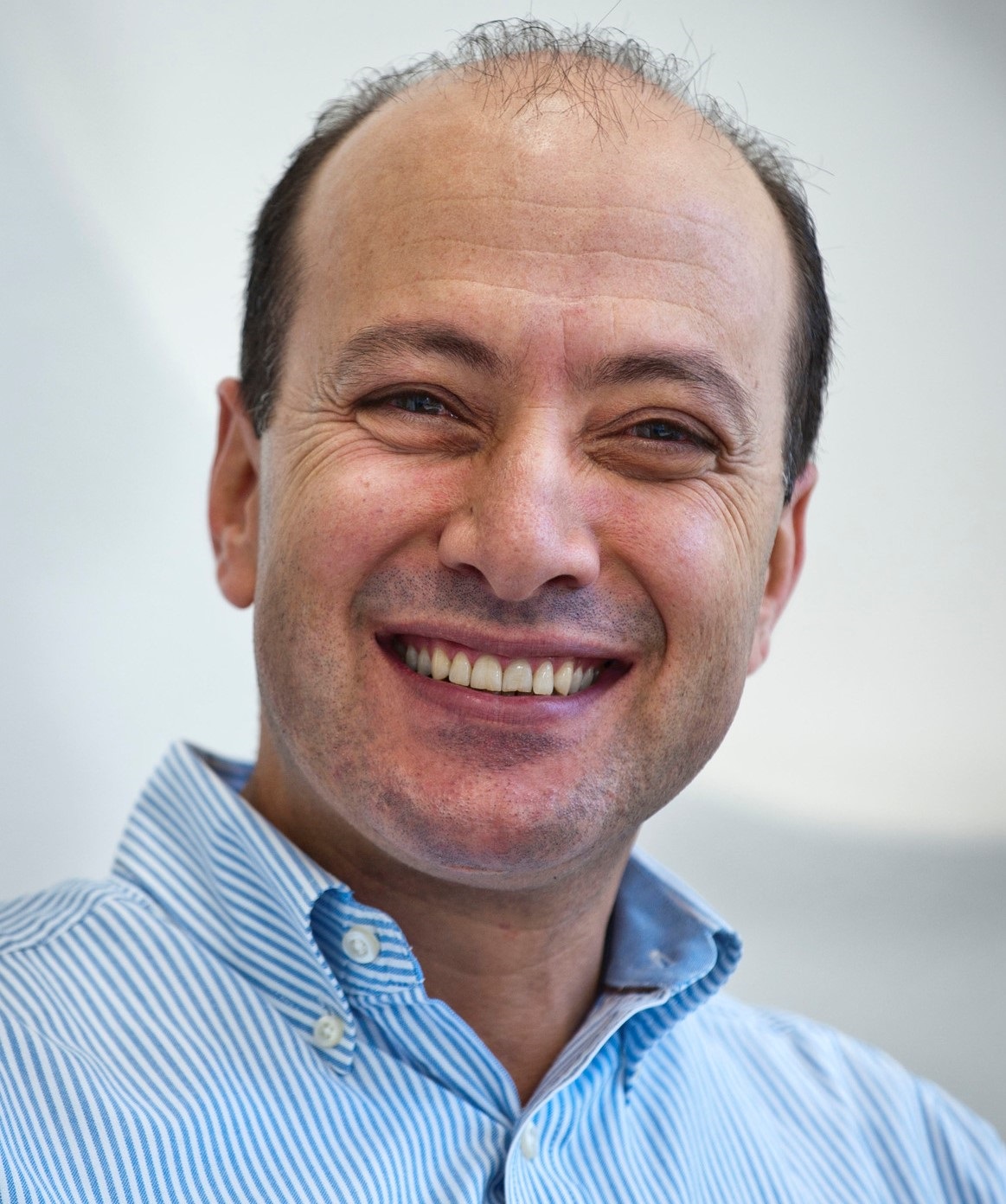}}]{Tamim Asfour} (Fellow, IEEE) is Full Professor of Humanoid Robotics at Karlsruhe Institute of Technology (KIT), Germany. He directs the High Performance Humanoid Technologies Lab (H2T) at the Institute of Anthropomatics and Robotics. He received a diploma in Electrical Engineering from the University of Karlsruhe and a Ph.D. degree from KIT. His research focuses on the engineering humanoid robot systems that integrates artificial intelligence, informatics, and mechatronics to perform versatile tasks in the real world while learning from humans, experience and interaction with the environment. Tamim is the developer of the ARMAR humanoid robot family. He has been a visiting professor at Georgia Tech, Tokyo University of Agriculture and Technology, and the National University of Singapore. He is a IEEE Fellow, Editor-in-Chief of the Robotics and Automation Letters (RA-L), Founding Editor-in-Chief of the IEEE-RAS Humanoids Conference Editorial Board, the Spokesperson of the Robotics Institute Germany (RIG), and the scientific spokesperson of the KIT Center “Information · Systems · Technologies” (KCIST).
\end{IEEEbiography}

\begin{IEEEbiography}[{\includegraphics[width=1in,clip,keepaspectratio]{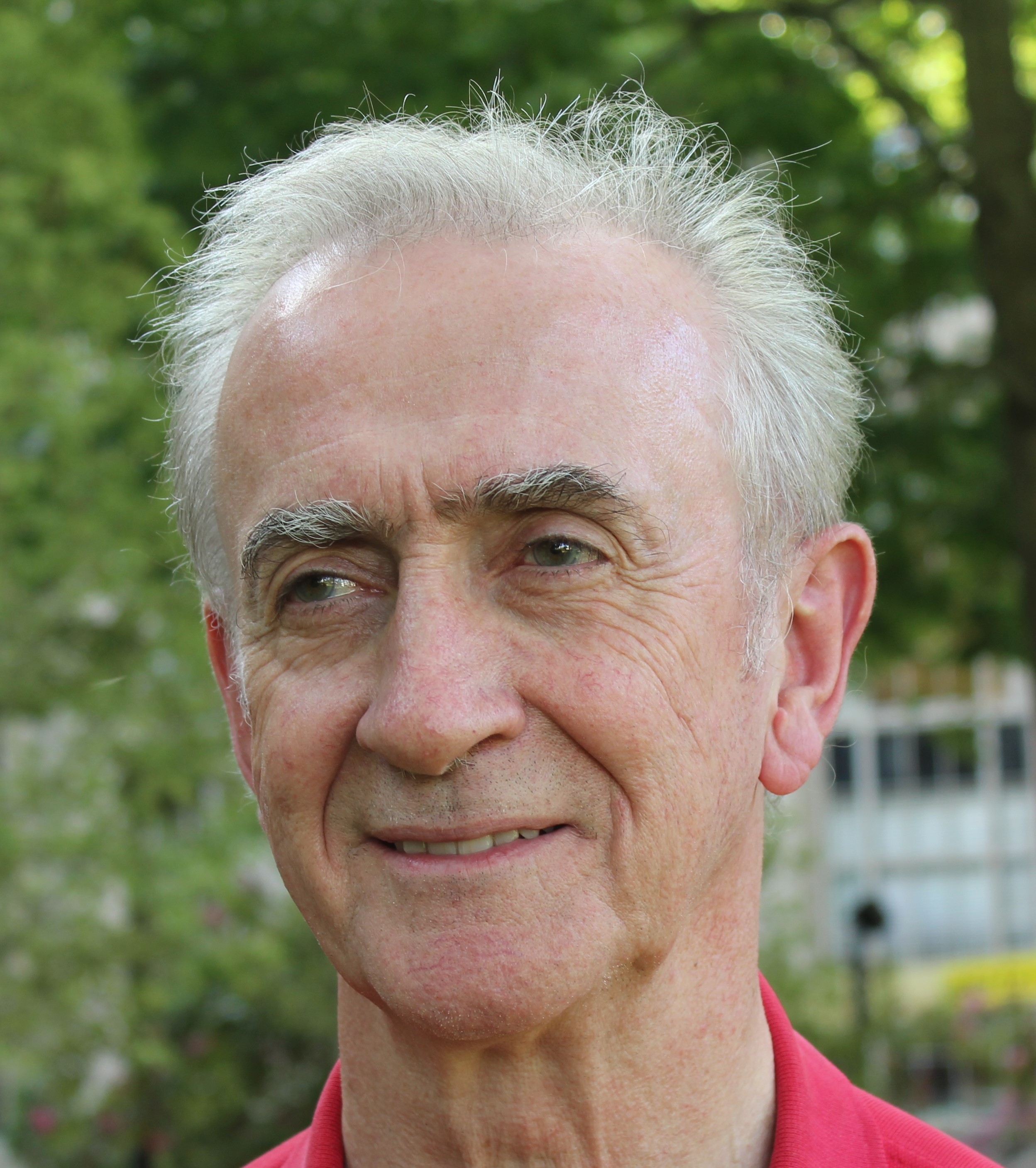}}]{Neville Hogan} (Member, IEEE) is Sun Jae Professor of Mechanical Engineering and Professor of Brain and Cognitive Sciences at the Massachusetts Institute of Technology. He earned a Diploma in Engineering (with distinction) from Dublin Institute of Technology and M.S., Mechanical Engineer and Ph.D. degrees from MIT. He joined MIT’s faculty in 1979 and presently directs the Newman Laboratory for Biomechanics and Human Rehabilitation. He co-founded Interactive Motion Technologies, subsequently part of Bionik Laboratories. His research includes robotics, motor neuroscience, and rehabilitation engineering, emphasizing the control of physical contact and dynamic interaction. Awards include: Honorary Doctorates from Delft University of Technology and Dublin Institute of Technology; the Silver Medal of the Royal Academy of Medicine in Ireland; the Saint Patrick’s Day Medal for Academia from Science Foundation Ireland; the Henry M. Paynter Outstanding Investigator Award and the Rufus T. Oldenburger Medal from the American Society of Mechanical Engineers, Dynamic Systems and Control Division; and from the Institute of Electrical and Electronics Engineers, the Academic Career Achievement Award from the Engineering in Medicine and Biology Society and the Pioneer in Robotics Award from the Robotics and Automation Society. 
\end{IEEEbiography}

\begin{IEEEbiography}[{\includegraphics[trim=10cm 0cm 6cm 0cm, clip, width=1in,height=1.25in,clip,keepaspectratio]{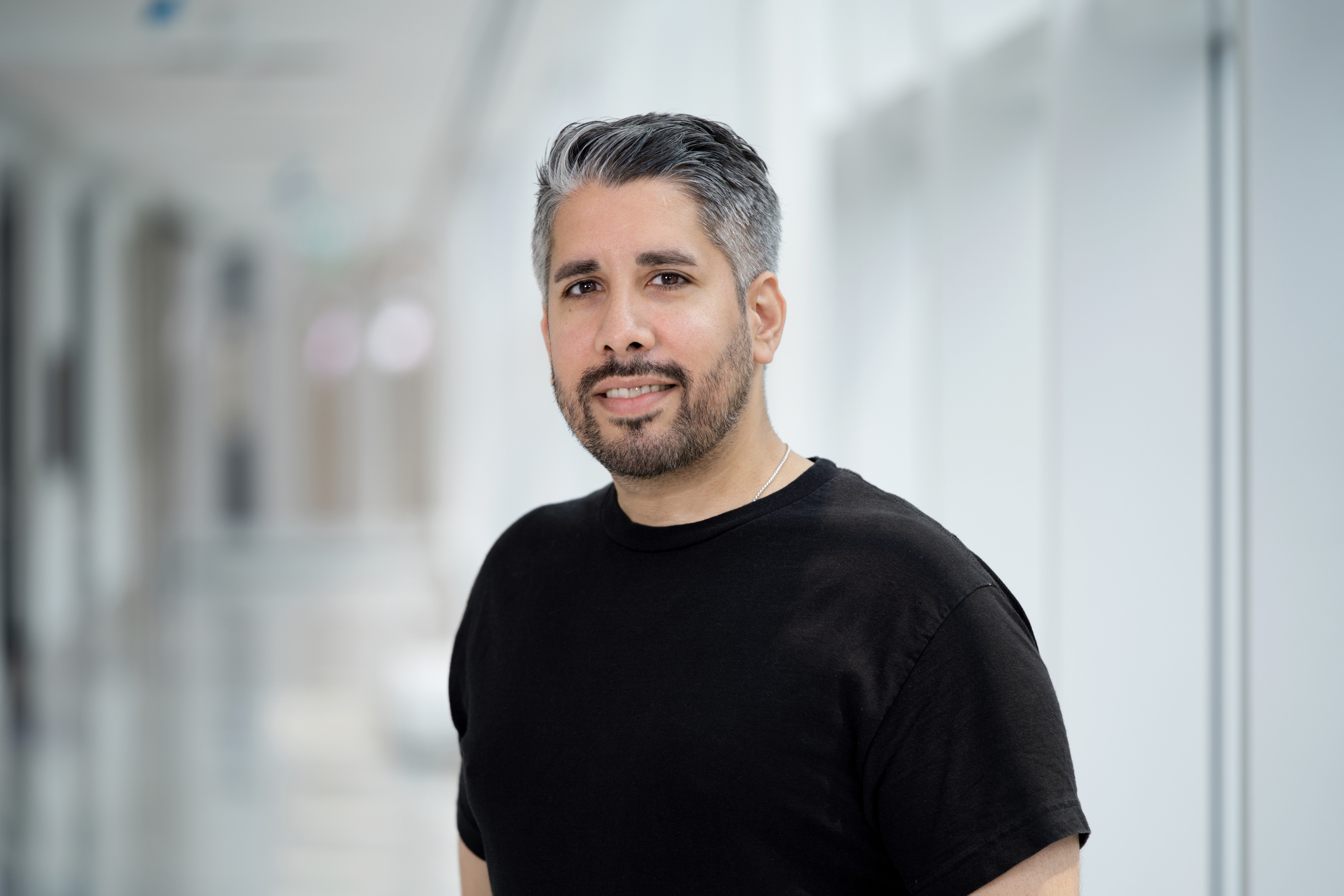}}]{Johannes Lachner}
(Member, IEEE) is an incoming Assistant Professor with the Elmore Family School of Electrical and Computer Engineering at Purdue University, West Lafayette, IN, USA. He is currently a Postdoctoral Researcher with the Department of Mechanical Engineering and the Department of Brain and Cognitive Sciences at the Massachusetts Institute of Technology (MIT). He received the M.Eng. double degree (with distinction) in Mechatronic Systems from the University of Ulster, Belfast, UK, and the University of Applied Sciences Augsburg, Germany, and the Ph.D. degree (with highest distinction) in Robotics from the University of Twente, Enschede, The Netherlands. Prior to MIT, he was a Senior Research Engineer with the Corporate Research Department, KUKA Robotics GmbH, where he invented numerous patents in robotics, several of which have been transferred into commercial products. His research focuses on rehabilitation robotics, motor neuroscience, and AI for physical human–robot interaction, with an emphasis on geometric methods for contact-rich robot control and robot learning from human physical interaction. He received the MIT–Novo Nordisk AI Fellowship Award, the IROS Best Paper Award 2024, and two consecutive TU Munich Global Incentive Fund awards. He was a finalist for the European Georges Giralt Ph.D. Award for the best European Ph.D. thesis in robotics.
\end{IEEEbiography}

\end{document}